\def\year{2022}\relax
\documentclass[letterpaper]{article} 
\usepackage{aaai22}  
\usepackage{times}  
\usepackage{helvet}  
\usepackage{courier}  
\usepackage[hyphens]{url}  
\usepackage{graphicx} 
\usepackage{natbib}  
\usepackage{caption} 
\DeclareCaptionStyle{ruled}{labelfont=normalfont,labelsep=colon,strut=off} 
\usepackage{rotating}
\usepackage{array}
\usepackage{algorithm}
\usepackage{algorithmic}

\usepackage{epsfig}
\usepackage{amsmath}
\usepackage{amssymb}
\usepackage{booktabs}
\usepackage{subfig}
\usepackage{paralist}
\usepackage{tabularx}
\usepackage{marvosym}
\usepackage{xcolor}
\usepackage{newfloat}
\usepackage{listings}
\floatstyle{ruled}
\newfloat{listing}{tb}{lst}{}
\floatname{listing}{Listing}
\def\our{PluGeN}

\def\R{\mathbb{R}}

\def\D{\mathcal{D}}
\def\Z{\mathcal{Z}}
\def\G{\mathcal{G}}
\def\F{\mathcal{F}}
\def\Z{\mathcal{Z}}
\def\N{\mathcal{N}}
\def\X{\mathcal{X}}

\title{\our{}: Multi-Label Conditional Generation From Pre-Trained Models}
\author{%
  \bf{Maciej Wo\l{}czyk}$^1$\thanks{Equal contribution},
  \bf{Magdalena Proszewska}$^1{}^*$,
  \bf{\L{}ukasz Maziarka}$^1$,\\
  \bf{Maciej Zieba}$^{2,  4}$,
  \bf{Patryk Wielopolski}$^2$,
  \bf{Rafa\l{} Kurczab}$^3$,
  \bf{Marek \'Smieja}$^1$\thanks{Corresponding author: \texttt{marek.smieja@uj.edu.pl}}
  }
\affiliations{
    $^1$Jagiellonian University, $^2$Wroclaw University of Science and Technology, $^3$ Institute of Pharmacology PAS, $^4$ Tooploox
}

\usepackage{bibentry}

\begin{document}

\maketitle

\begin{abstract}
Modern generative models achieve excellent quality in a variety of tasks including image or text generation and chemical molecule modeling. However, existing methods often lack the essential ability to generate examples with requested properties, such as the age of the person in the photo or the weight of the generated molecule. Incorporating such additional conditioning factors would require rebuilding the entire architecture and optimizing the parameters from scratch. Moreover, it is difficult to disentangle selected attributes so that to perform edits of only one attribute while leaving the others unchanged. To overcome these limitations we propose \our{} (Plugin Generative Network), a simple yet effective generative technique that can be used as a plugin to pre-trained generative models. The idea behind our approach is to transform the entangled latent representation using a flow-based module into a multi-dimensional space where the values of each attribute are modeled as an independent one-dimensional distribution. In consequence, \our{} can generate new samples with desired attributes as well as manipulate labeled attributes of existing examples. Due to the disentangling of the latent representation, we are even able to generate samples with rare or unseen combinations of attributes in the dataset, such as a young person with gray hair, men with make-up, or women with beards. We combined \our{} with GAN and VAE models and applied it to conditional generation and manipulation of images and chemical molecule modeling. Experiments demonstrate that \our{} preserves the quality of backbone models while adding the ability to control the values of labeled attributes. Implementation is available at \texttt{https://github.com/gmum/plugen}.
\end{abstract}

\section{Introduction}

\newcolumntype{C}{>{\centering\arraybackslash}X}
\begin{figure}[ht]
    \centering
    \begin{tabularx}{\linewidth}{CCCC} input & gender & \hspace*{-2mm} glasses & smile\end{tabularx}\\
    \includegraphics[width=\linewidth]{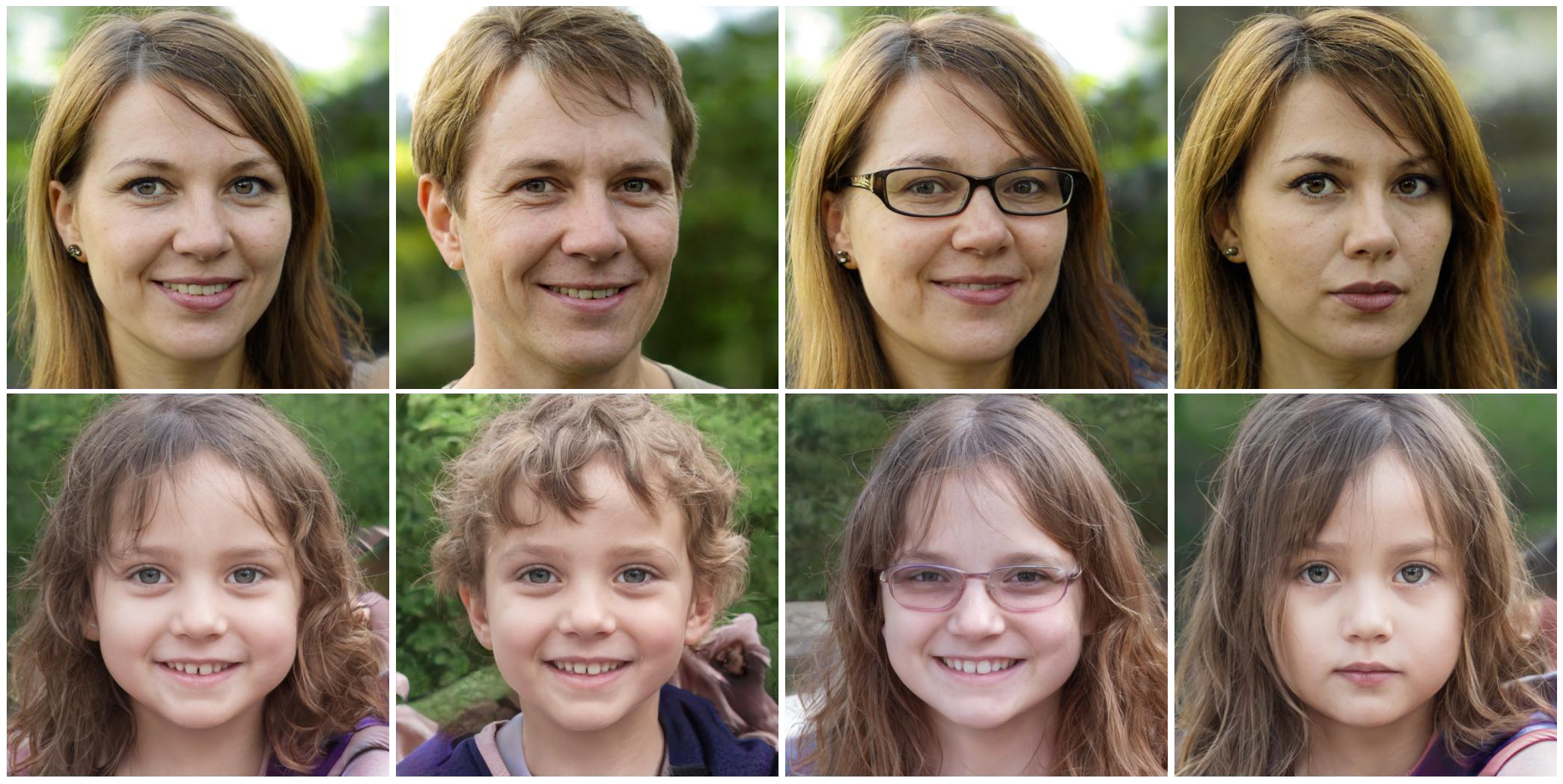}
    \caption{Attributes manipulation performed by \our{} using the StyleGAN backbone.}
    \label{fig:change-stylegan}
\end{figure}

Generative models such as GANs and variational autoencoders have achieved great results in recent years, especially in the domains of images \cite{brock2018large,DBLP:conf/nips/BrownMRSKDNSSAA20} and cheminformatics \cite{gomez2018automatic,jin2018junction}. 
However, in many practical applications, we need to control the process of creating samples by enforcing particular features of generated objects. This would be required to regulate the biases present in the data, e.g. to assure that people of each ethnicity are properly represented in the generated set of face images. In numerous realistic problems, such as drug discovery, we want to find objects with desired properties, like molecules with a particular activity, non-toxicity, and solubility.

\begin{figure*}
    \centering
     \subfloat[Factorization of true data distribution\label{fig:plugen64_label_change}]{\includegraphics[width=0.4\linewidth]{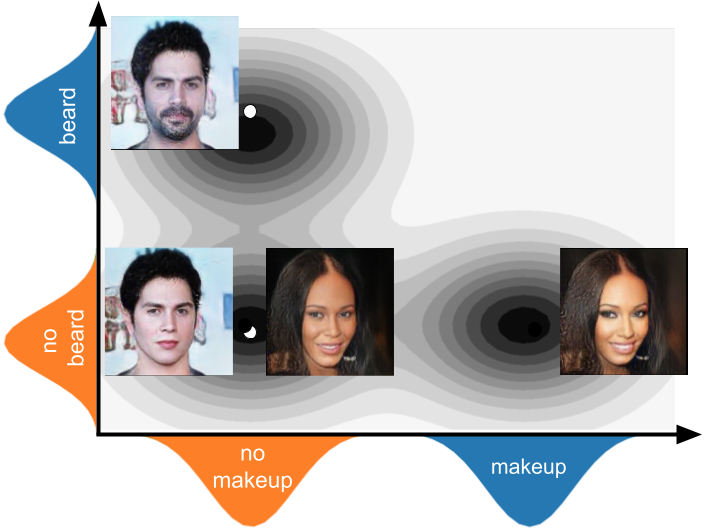}} \hspace{1cm}
    \subfloat[Probability distribution covered by \our{}.\label{fig:semi_plugen64_label_change}]{\includegraphics[width=0.4\linewidth]{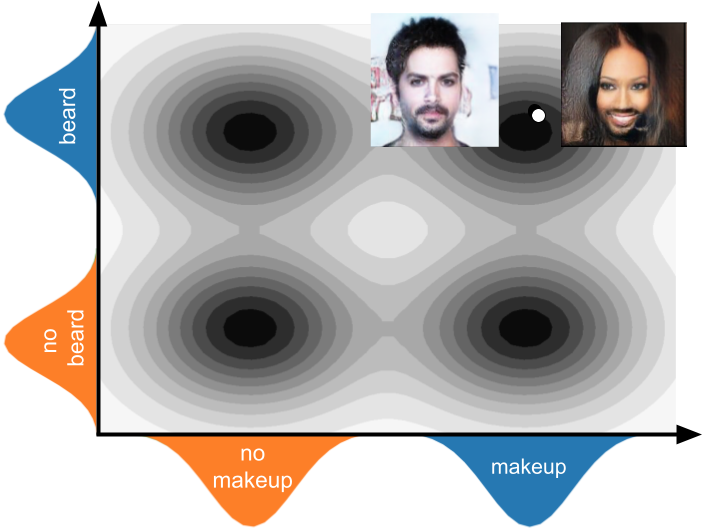}}
  \caption{\our{} factorizes true data distribution into components (marginal distributions) related to labeled attributes, see (a), and allows for describing unexplored regions of data (uncommon combinations of labels) by sampling from independent components, see (b). In the case illustrated here, \our{} constructs pictures of men with make-up or women with beards, although such examples rarely (or never) appear in the training set.}
    \label{fig:teaser}
\end{figure*}

Designing the conditional variants of generative models that operate on multiple labels is a challenging problem due to intricate relations among the attributes. Practically, it means that some combinations of attributes (e.g. a woman with a beard) might be unobserved or rarely observed in the training data. In essence, the model should be able to go beyond the distribution of seen data and generate examples with combinations of attributes not encountered previously. One might approach this problem by building a new conditional generative model from the ground up or design a solution tailored for a specific existing unsupervised generative model. However, this introduces an additional effort when one wants to adapt it to a newly invented approach. 

To tackle this problem while leveraging the power of existing techniques, we propose \our{} (Plugin Generative Network), a simple yet effective generative technique that can be used as a plugin to various pre-trained generative models such as VAEs or GANs, see Figure \ref{fig:change-stylegan} for demonstration. Making use of \our{}, we can manipulate the attributes of input examples as well as generate new samples with desired features. When training the proposed module, we do not change the parameters of the base model and thus we retain its generative and reconstructive abilities, which places our work in the emerging family of non-invasive network adaptation methods \cite{wolczyk2021zero, DBLP:conf/nips/RebuffiBV17,koperski2020plugin,DBLP:conf/icml/KayaHD19,DBLP:conf/nips/ZhouXGM0W20}.

Our idea is to find a mapping between the entangled latent representation of the backbone model and a disentangled space, where each dimension corresponds to a single, interpretable attribute of the image.  By factorizing the true data distribution into independent components, we can sample from each component independently, which results in creating samples with arbitrary combinations of attributes, see Figure \ref{fig:teaser}. In contrast to many previous works, which are constrained to the attributes combinations visible in the training set, \our{} gives us full control of the generation process, being able to create uncommon combinations of attributes, such as a woman with a beard or a man with heavy make-up. Generating samples with unseen combinations of attributes can be viewed as extending the distribution of generative models to unexplored although reasonable regions of data space, which distinguishes our approach from existing solutions.

Extensive experiments performed on the domain of images and a dataset of chemical compounds demonstrate that \our{} is a reusable plugin that can be applied to various architectures including GANs and VAEs. In contrast to the baselines, \our{} can generate new samples as well as manipulate the properties of existing examples, being capable of creating uncommon combinations of attributes.

Our contributions are as follow:
\begin{compactitem}
    \item We propose a universal and reusable plugin for multi-label generation and manipulation that can be attached to various generative models and applied it to diverse domains, such as chemical molecule modeling.
    \item We introduce a novel way of modeling conditional distributions using invertible normalizing flows based on the latent space factorization.
    \item We experimentally demonstrate that \our{} can produce samples with uncommon combinations of attributes going beyond the distribution of training data.
\end{compactitem}

\section{Related work}

Conditional VAE (cVAE) is one of the first methods which includes additional information about the labeled attributes in a generative model~\cite{kingma2014semi}. Although this approach has been widely used in various areas ranging from image generation~\cite{sohn2015learning,yan2016attribute2image,klys2018learning} to molecular design~\cite{kang2018conditional}, the independence of the latent vector from the attribute data is not assured, which negatively influences the generation quality.
Conditional GAN (cGAN) is an alternative approach that gives results of significantly better quality~\cite{mirza2014conditional,perarnau2016invertible,he2019attgan}, but the model is more difficult to train~\cite{kodali2017convergence}. cGAN works very well for generating new images and conditioning factors may take various forms (images, sketches, labels) \cite{park2019semantic, jo2019sc, choi2020stargan}, but manipulating existing examples is more problematic because GAN models lack the encoder network \cite{tov2021designing}. Fader Networks~\cite{lample2017fader} combine features of both cVAE and cGAN, as they use encoder-decoder architecture, together with the discriminator, which predicts the image attributes from its latent vector returned from the encoder. As discussed in \cite{li2020latent}, the training of Fader Networks is even more difficult than standard GANs, and disentanglement of attributes is not preserved. MSP \cite{li2020latent} is a recent auto-encoder based architecture with an additional projection matrix, which is responsible for disentangling the latent space and separating the attribute information from other characteristic information. In contrast to \our{}, MSP cannot be used with pre-trained GANs and performs poorly at generating new images (it was designed for manipulating existing examples). CAGlow~\cite{liu2019conditional} is an adaptation of Glow~\cite{kingma2018glow} to conditional image generation based on modeling a joint probabilistic density of an image and its conditions. Since CAGlow does not reduce data dimension, applying it to more complex data might be problematic. 

While the above approaches focus on building conditional generative models from scratch, recent works often focus on manipulating the latent codes of pre-trained models. StyleFlow~\cite{abdal2020styleflow} operates on the latent space of StyleGAN~\cite{karras2019style} using a conditional continuous flow module. Although the quality of generated images is impressive, the model has not been applied to other generative models than StyleGAN and domains other than images. Moreover, StyleFlow needs an additional classifier to compute the conditioning factor (labels) for images at test time. Competitive approaches to StyleGAN appear in \cite{gao2021high, tewari2020pie, harkonen2020ganspace, nitzan2020disentangling}. InterFaceGAN \cite{shen2020interfacegan} postulates that various properties of the facial semantics can be manipulated via linear models applied to the latent space of GANs. Hijack-GAN~\cite{wang2020hijack} goes beyond linear models and designs a proxy model to traverse the latent space of GANs.

In disentanglement learning, we assume that the data has been generated from a fixed number of independent factors of underlying variation. The goal is then to find a transformation that unravels these factors so that a change in one dimension of the latent space corresponds to a change in one factor of variation while being relatively invariant to changes in other factors~\cite{bengio2013representation,kim2018disentangling,higgins2016beta,brakel2017learning,kumar2017variational,chen2019isolating,spurek2020non,dinh2014nice,sorrenson2020disentanglement,chen2016infogan}. As theoretically shown in \cite{locatello2019challenging}, the unsupervised learning of disentangled representations is fundamentally impossible without inductive biases on both the models and the data. In this paper, we solve a slightly different problem than typical disentanglement, as we aim to deliver an efficient plug-in model to a large variety of existing models in order to manipulate attributes without training the entire system. Creating compact add-ons for large models saves training time and energy consumption.

\section{Plugin Generative Network}

\begin{figure}[t!]
    \centering
    \includegraphics[width=0.48\textwidth]{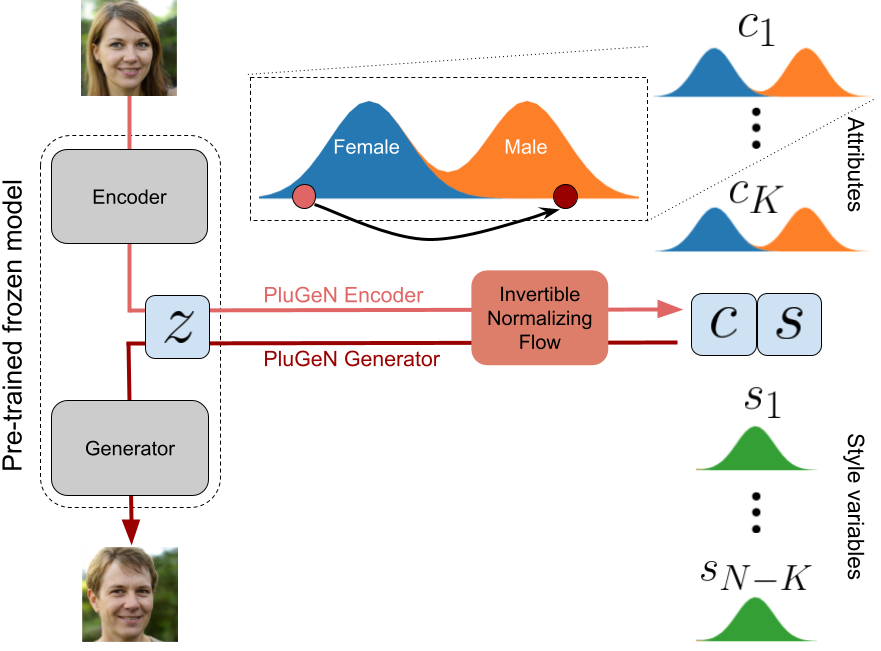}
    \caption{\our{} maps the entangled latent space $\Z$ of pre-trained generative models using invertible normalizing flow into a separate space, where labeled attributes are modeled using independent 1-dimensional distributions. By manipulating label variables in this space, we fully control the generation process. }
    \label{fig:scheme}
\end{figure}

We propose a plugin generative network (\our), which can be attached to pre-trained generative models and allows for direct manipulation of labeled attributes, see Figure \ref{fig:scheme} for the basic scheme of \our{}. Making use of \our{} we preserve all properties of the base model, such as generation quality and reconstruction in the case of auto-encoders, while adding new functionalities. In particular, we can:
\begin{compactitem}
\item modify selected attributes of existing examples,
\item generate new samples with desired labels.
\end{compactitem}
In contrast to typical conditional generative models, \our{} is capable of creating examples with rare or even unseen combinations of attributes, e.g. man with makeup.

{\bf Probabilistic model.} \our{} works in a multi-label setting, where every example $\mathbf{x} \in \X$ is associated with a $K$-dimensional vector of binary labels\footnote{Our model can be extended to continuous values, which we describe in the supplementary materials due to page limit.} $\mathbf{y} = (y_1,\ldots,y_K) \in \{0, 1\}^K$. 
We assume that there is a pre-trained generative model $\G: \Z \to \R^D$, where $\Z \subset \R^N$ is the latent space, which is usually heavily entangled. That is, although each latent code $\mathbf{z} \in \Z$ contains the information about the labels $\mathbf{y}$, there is no direct way to extract or modify it. 

We want to map this entangled latent space $\Z$ into a separate latent space $\D \subset \R^N$ which encodes the values of each label $y_k$ as a separate random variable $C_k$ living in a single dimension of this space. Thus, by changing the value of $C_k$, going back to the entangled space $\Z$ and generating a sample, we can control the values of $y_k$. Since labeled attributes usually do not fully describe a given example, we consider additional $N-K$ random variables $S_k$, which are supposed to encode the information not included in the labels. We call $\mathbf{C} = (C_1,\ldots,C_K)$ the label variables (or attributes) and $\mathbf{S}=(S_1,\ldots,S_{N-K})$ the style variables.

Since we want to control the value of each attribute independently of any other factors, we assume the factorized form of the probability distribution of the random vector $(\mathbf{C},\mathbf{S})$. More precisely, the conditional probability distribution of $(\mathbf{C},\mathbf{S})$ given any condition $\mathbf{Y}=\mathbf{y}$ imposed on labeled attributes is of the form:
\begin{equation} \label{eq:factorized}
p_{\mathbf{C},\mathbf{S}|\mathbf{Y}=\mathbf{y}}(\mathbf{c},\mathbf{s}) = \prod_{i=1}^K p_{C_i|Y_i=y_i}(c_i) \cdot p_{\mathbf{S}}(\mathbf{s}),
\end{equation}
for all $(\mathbf{c},\mathbf{s}) = (c_1,\ldots,c_K,s_1,\ldots,s_{N-K}) \in \R^N$. In other words, modifying $Y_i = y_i$ influences only the $i$-th factor $C_i$ leaving other features unchanged. 

\textbf{Parametrization.} To instantiate the above probabilistic model \eqref{eq:factorized}, we need to parametrize the conditional distribution of $C_i$ given $Y_i=y_i$ and the distribution of $\mathbf{S}$. Since we do not impose any constraints on style variables, we use standard Gaussian distribution for modeling density of $\mathbf{S}$:
$$
p_{\mathbf{S}} = \N(0,I_{N-K}).
$$

To provide the consistency with $p_{\mathbf{S}}$ and avoid potential problems with training our deep learning model using discrete distributions, we use the mixture of two Gaussians for modeling the presence of labels -- each component corresponds to a potential value of the label ($0$ or $1$). More precisely, the conditional distribution of $C_i$ given $Y_i=y_i$ is parametrized by:
\begin{equation}\label{eq:condG}
p_{C_i|Y_i=y_i} = \N(m_0, \sigma_0)^{(1-y_i)} \cdot \N(m_1, \sigma_1)^{y_i},
\end{equation}
where $m_0,m_1,\sigma_0,\sigma_1$ are the user-defined parameters. If $y_i=0$, then the latent factor $C_i$ takes values close to $m_0$; otherwise we get values around $m_1$ (depending on the value of $\sigma_0$ and $\sigma_1$). To provide good separation between components, we put $m_0 = -1, m_1 = 1$; the selection of $\sigma_0,\sigma_1$ will be discussed is the supplementary materials.

Thanks to this continuous parametrization, we can smoothly interpolate between different labels, which would not be so easy using e.g. Gumbel softmax parametrization \cite{jang2016categorical}. In consequence, we can gradually change the intensity of certain labels, like smile or beard, even though such information was not available in a training set (see Figure \ref{fig:interpolation-stylegan} in the experimental section).

\paragraph{Training the model}

To establish a two-way mapping between entangled space $\Z$ and the disentangled space $\D$, we use an invertible normalizing flow (INF), $\F: \R^N \to \Z$. Let us recall that INF is a neural network, where the inverse mapping is given explicitly and the Jacobian determinant can be easily calculated \cite{dinh2014nice}. Due to the invertibility of INF, we can transform latent codes $\mathbf{z} \in \Z$ to the prior distribution of INF, modify selected attributes, and map the resulting vector back to $\Z$. Moreover, INFs can be trained using log-likelihood loss, which is very appealing in generative modeling. 

Summarizing, given a latent representation $\mathbf{z} \in \Z$ of a sample $\mathbf{x}$ with label $\mathbf{y}$, the loss function of \our{} equals:
\begin{multline} \label{eq:loss}
-\log p_{\mathbf{Z}|\mathbf{Y}= \mathbf{y}}(\mathbf{z}) =\\
-\log \left(p_{\mathbf{C},\mathbf{S} | \mathbf{Y} = \mathbf{y}}(\mathbf{c},\mathbf{s}) \cdot \left|\det \frac{\partial \F^{-1}(\mathbf{z})}{\partial \mathbf{z}}\right| \right)=\\
 -\log \left( \prod_{i=1}^K p_{C_i|Y_i=y_i}(c_i) \cdot p_{\mathbf{S}}(\mathbf{s}) \right) - \log \left|\det \frac{\partial \F^{-1}(\mathbf{z})}{\partial \mathbf{z}}\right|=\\
 -\sum_{i=1}^K \log p_{C_i|Y_i=y_i}(c_i) - \log p_{\mathbf{S}}(\mathbf{s}) - \log \left|\det \frac{\partial \F^{-1}(\mathbf{z})}{\partial \mathbf{z}}\right|,
\end{multline}
where $(\mathbf{c},\mathbf{s}) = \F^{-1}(\mathbf{z})$.
In the training phase, we collect latent representations $\mathbf{z}$ of data points $\mathbf{x}$. Making use of labeled attributes $\mathbf{y}$ associated with every $\mathbf{x}$, we modify the weights of $\F$ so that to minimize the negative log-likelihood \eqref{eq:loss} using gradient descent. The weights of the base model $\G$ are kept frozen. 

In contrast to many previous works \cite{abdal2020styleflow}, \our{} can be trained in a semi-supervised setting, where only partial information about labeled attributes is available (see supplementary materials for details).

{\bf Inference.} We may use \our{} to generate new samples with desired attributes as well as to manipulate attributes of input examples. In the first case, we generate a vector $(\mathbf{c},\mathbf{s})$ from the conditional distribution $p_{\mathbf{C},\mathbf{S} | \mathbf{Y}=\mathbf{y}}$ with selected condition $\mathbf{y}$. To get the output sample, the vector $(\mathbf{c},\mathbf{s})$ is transformed by the INF and the base generative network $\G$, which gives us the final output $\mathbf{x}=\G(\F(\mathbf{c},\mathbf{s}))$. 

In the second case, to manipulate the attributes of an existing example $\mathbf{x}$, we need to find its latent representation $\mathbf{z}$. If $\G$ is a decoder network of an autoencoder model, then $\mathbf{x}$ should be passed through the encoder network to obtain $\mathbf{z}$ \cite{li2020latent}. If $\G$ is a GAN, then $\mathbf{z}$ can be found by minimizing the reconstruction error between $\mathbf{x'}=\G(\mathbf{z})$ and $\mathbf{x}$ using gradient descent for a frozen $\G$ \cite{abdal2020styleflow}. In both cases, $\mathbf{z}$ is next processed by INF, which gives us its factorized representation $(\mathbf{c}, \mathbf{s}) = \F^{-1}(\mathbf{z})$. In this representation, we can modify any labeled variable $c_i$ and map the resulting vector back through $\F$ and $\G$ as in the generative case.

Observe that \our{} does not need to know what are the values of labeled attributes when it modifies attributes of existing examples. Given a latent representation $\mathbf{z}$, \our{} maps it through $\G^{-1}$, which gives us the factorization into labeled and unlabeled attributes. In contrast, existing solutions based on conditional INF, e.g StyleFlow \cite{abdal2020styleflow}, have to determine all labels before passing $\mathbf{z}$ through INF as they represent the conditioning factors. In consequence, these models involve additional classifiers for labeled attributes.

\section{Experiments}

To empirically evaluate the properties of \our{}, we combine it with GAN and VAE architectures to manipulate attributes of image data. Moreover, we present a practical use-case of chemical molecule modeling using CharVAE. Due to the page limit, we included architecture details and additional results in the supplementary materials.

\paragraph{GAN backbone} First, we consider the state-of-the-art StyleGAN architecture \cite{karras2019style}, which was trained on Flickr-Faces-HQ (FFHQ) containing 70 000 high-quality images of resolution $1024 \times 1024$. The Microsoft Face API was used to label 8 attributes in each image (gender, pitch, yaw, eyeglasses, age, facial hair, expression, and baldness). 

\our{} is instantiated using NICE flow model \cite{dinh2014nice} that operates on the latent vectors $\mathbf{w} \in \R^{512}$ sampled from the $\mathbf{W}$ space of the StyleGAN. As a baseline, we select StyleFlow \cite{abdal2020styleflow}, which is currently one of the state-of-the-art models for controlling the generation process of StyleGAN. In contrast to \our{}, StyleFlow uses the conditional continuous INF to operate on the latent codes of StyleGAN, where the conditioning factor corresponds to the labeled attributes. 
For evaluation, we modify one of 5 attributes\footnote{The remaining 3 attributes (age, pitch, yaw) are continuous and it is more difficult to assess their modifications.} and verify the success of this operation using the prediction accuracy returned by Microsoft Face API. The quality of images is additionally assessed by calculating the standard Fréchet Inception Distance (FID) \cite{DBLP:conf/nips/HeuselRUNH17}.

\begin{figure*}[ht]
    \centering
    \subfloat[\our{}]{\includegraphics[width=0.48\linewidth]{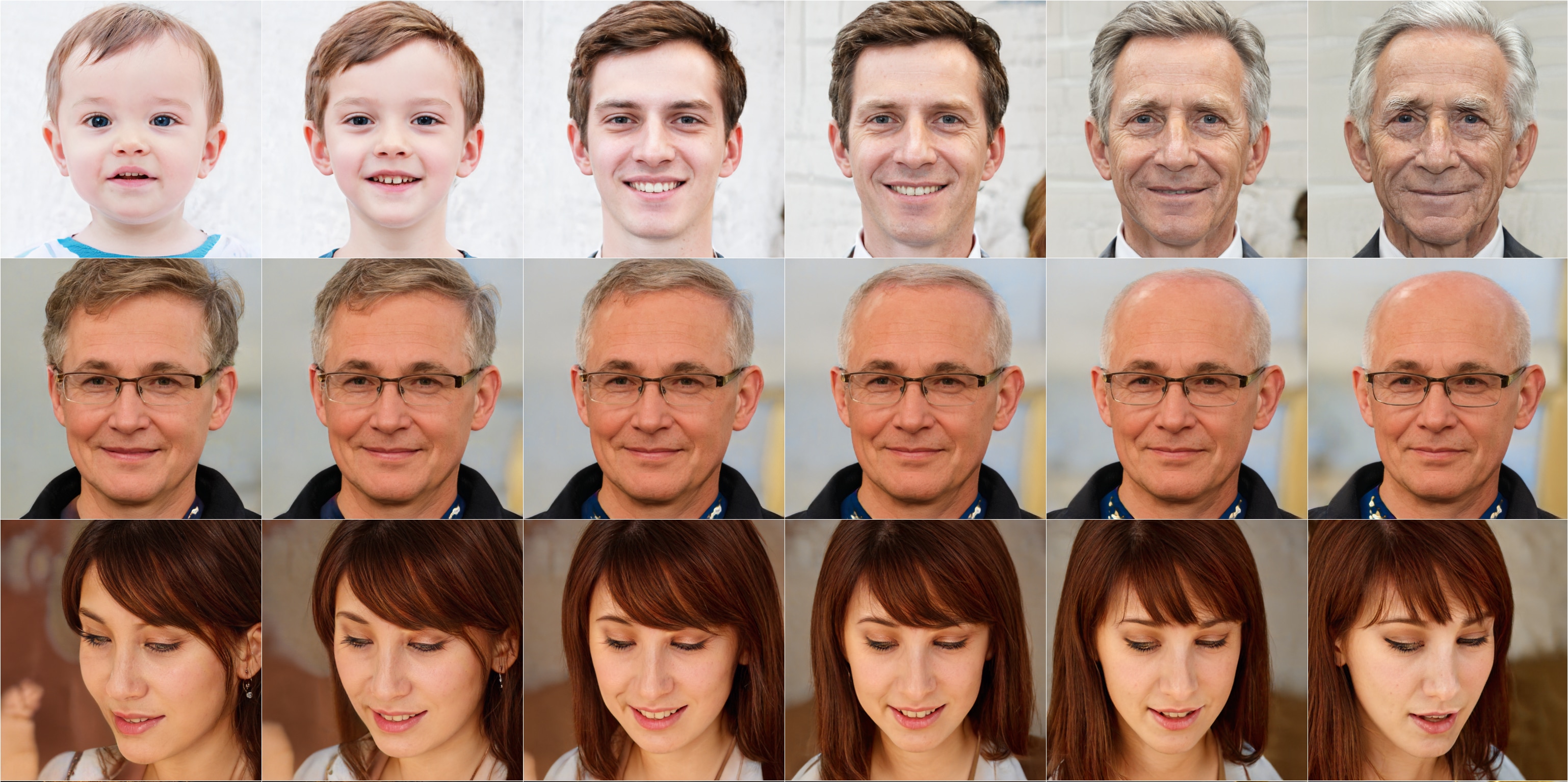}} \quad
    \subfloat[StyleFlow]{\includegraphics[width=0.48\linewidth]{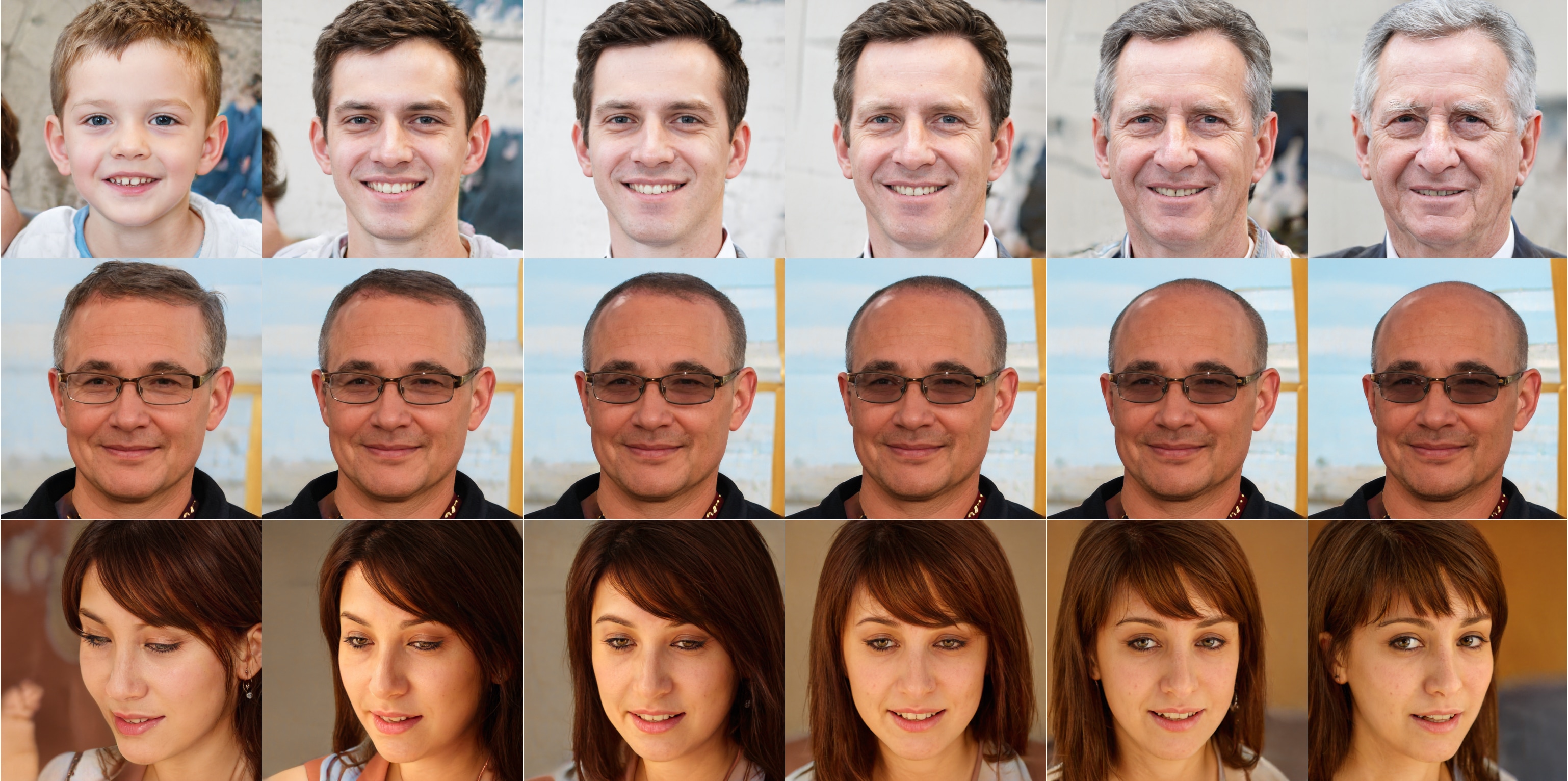}}
    \caption{Gradual modification of attributes (age, baldness, and yaw, respectively) performed on the StyleGAN latent codes.}
    \label{fig:interpolation-stylegan}
\end{figure*}

Figure \ref{fig:change-stylegan} (first page) and \ref{fig:interpolation-stylegan} present the effects of how \our{} and StyleFlow manipulate images sampled by StyleGAN. It is evident that \our{} can switch the labels to opposite values as well as gradually change their intensities. At the same time, the requested modifications do not influence the remaining attributes leaving them unchanged. One can observe that the results produced by StyleFlow are also acceptable, but the modification of the requested attribute implies the change of other attributes. For example, increasing the intensity of "baldness" changes the type of glasses, or turning the head into right makes the woman look straight.

The above qualitative evaluation is supported by the quantitative assessment presented in Table \ref{tab:stylegan}. 
As can be seen, StyleFlow obtains a better FID score, while \our{} outperforms StyleFlow in terms of accuracy. Since FID compares the distribution of generated and real images, creating images with uncommon combinations of attributes that do not appear in a training set may be scored lower, which can explain the relation between accuracy and FID obtained by \our{} and StyleFlow. In consequence, FID is not an adequate metric for measuring the quality of arbitrary image manipulations considered here, because it is too closely tied to the distribution of input images. 

It is worth mentioning that \our{} obtains these very good results using NICE model, which is the simplest type of INFs. In contrast, StyleFlow uses continuous INF, which is significantly more complex and requires using an ODE solver leading to unstable training. Moreover, to modify even a single attribute, StyleFlow needs to determine the values of all labels, since they represent the conditioning factors of INF. In consequence, every modification requires applying an auxiliary classifier to predict all image labels. The usage of \our{} is significantly simpler, as subsequent coordinates in the latent space of INF correspond to the labeled attributes and they are automatically determined by \our{}. Finally, our approach is less computationally expensive as we verified that, using the same hardware, \our{} can be trained 3 times faster than StyleFlow and is around 100 times faster in inference.

\begin{table}[tb]
\caption{Accuracy and FID scores of attributes modification using StyleGAN backbone.} \label{tab:stylegan}
\centering
\begin{tabular}{lcc}
\toprule
Requested value  & \our{} & StyleFlow \\
\midrule
female       & {\bf 0.95} & {\bf 0.95}  \\
male         & {\bf 0.92} & 0.87  \\
no-glasses    & {\bf 1.00} & 0.99  \\
glasses      & {\bf 0.90} & 0.70  \\
not-bald      & {\bf 1.00} & {\bf 1.00}  \\
bald         & 0.53 & {\bf 0.54}  \\
no-facial-hair & {\bf 1.00} & {\bf 1.00}  \\
facial-hair   & {\bf 0.72} & 0.65  \\
no-smile      & {\bf 0.99} & 0.92  \\
smile        & 0.96 & {\bf 0.99}  \\
\midrule
Average Acc & {\bf 0.90} & 0.86 \\
Average FID      & 46.51  & {\bf 32.59} \\
\bottomrule           
\end{tabular}
\end{table}

\paragraph{Image manipulation on VAE backbone}

In the following experiment, we show that \our{} can be combined with autoencoder models to effectively manipulate image attributes. We use CelebA database, where every image of the size $256 \times 256$ is annotated with $40$ binary labels. 

We compare \our{} to MSP \cite{li2020latent}, a strong baseline, which uses a specific loss for disentangling the latent space of VAE. Following the idea of StyleFlow, we also consider a conditional INF attached to the latent space of pre-trained VAE (referred to as cFlow), where conditioning factors correspond to the labeled attributes. The architecture of the base VAE and the evaluation protocol were taken from the original MSP paper. More precisely, for every input image, we manipulate the values of two attributes (we inspect 20 combinations in total). The success of the requested manipulation is verified using a multi-label ResNet-56 classifier trained on the original CelebA dataset.

The sample results presented in Figure  \ref{fig:modifications} demonstrate that \our{} attached to VAE produces high-quality images satisfying the constraints imposed on the labeled attributes. The quantitative comparison shown in Table \ref{tab:image_manipulation} confirms that \our{} is extremely efficient in creating uncommon combinations of attributes, while cFlow performs well only for the usual combinations. At the same time, the quality of images produced by \our{} and MSP is better than in the case of cFlow. Although both \our{} and MSP focus on disentangling the latent space of the base model, MSP has to be trained jointly with the base VAE model and it was designed only to autoencoder models. In contrast, \our{} is a separate module, which can be attached to arbitrary pre-trained models. Due to the use of invertible neural networks, it preserves the reconstruction quality of the base model, while adding manipulation functionalities. In the following experiment, we show that \our{} also performs well at generating entirely new images, which is not possible using MSP.

\definecolor{gold}{rgb}{0.85,.66,0}
\definecolor{grey}{rgb}{0.8, 0.8, 0.8}
\begin{figure*}[ht]
    \subfloat{
    \setlength{\tabcolsep}{0pt}
    \begin{tabularx}{\linewidth}{cp{4pt}cp{4pt}c}
        {\begin{tabularx}{0.32\textwidth}{llC}
            \includegraphics[width=0.065\textwidth]{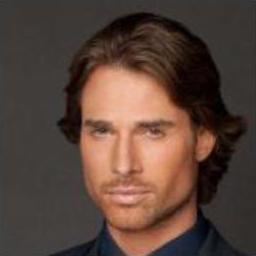} & \hspace{0.05cm} {\begin{turn}{90} \scriptsize Input image\end{turn}} & \vspace{-0.7cm} \hspace{-0.8cm} \our{}
        \end{tabularx}} & & MSP & & cFlow \\ \vspace{-2mm}
        \includegraphics[width=0.32\textwidth]{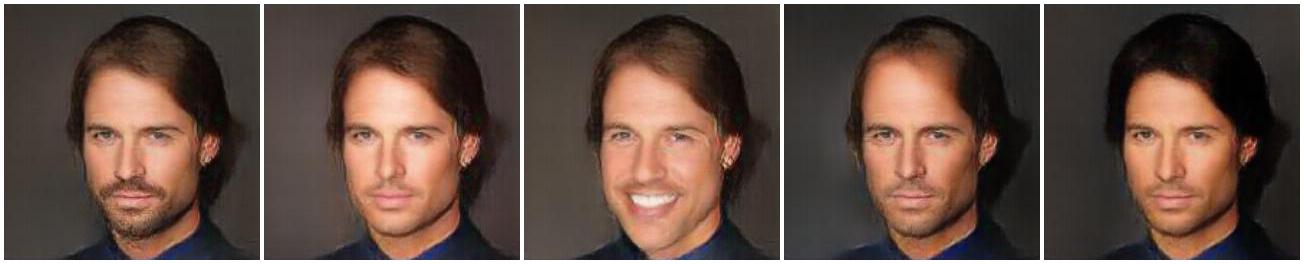} & & \includegraphics[width=0.32\textwidth]{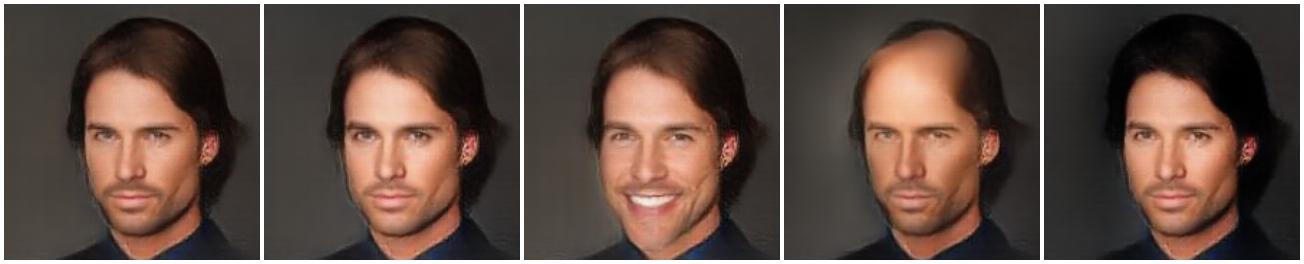} & & \includegraphics[width=0.32\textwidth]{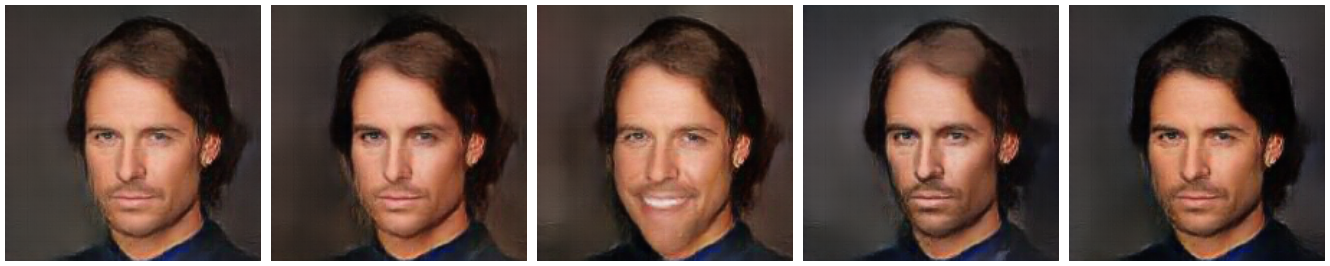} \\
        {\scriptsize \begin{tabularx}{0.32\textwidth}{CCCCC}\Male +beard & \Male +mkup & open+smile & \Male +bald & hair-glass \\ \end{tabularx}} & & {\scriptsize \begin{tabularx}{0.32\textwidth}{CCCCC}\Male +beard & \Male +mkup & open+smile & \Male +bald & hair-glass \\  \end{tabularx}} & & {\scriptsize \begin{tabularx}{0.32\textwidth}{CCCCC}\Male +beard & \Male +mkup & open+smile & \Male +bald & hair-glass \\ \end{tabularx}} \\ \vspace{-2mm}
        \includegraphics[width=0.32\textwidth]{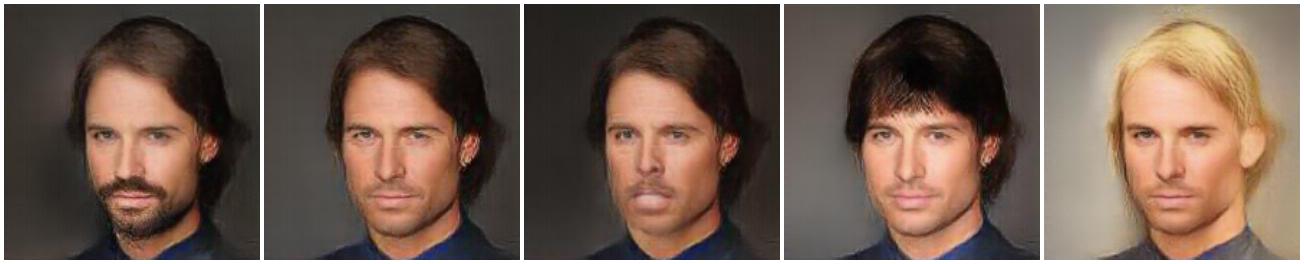} & & \includegraphics[width=0.32\textwidth]{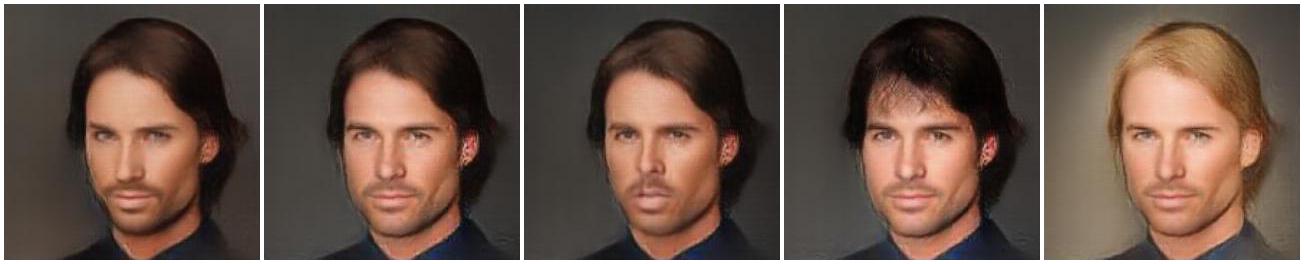} & & \includegraphics[width=0.32\textwidth]{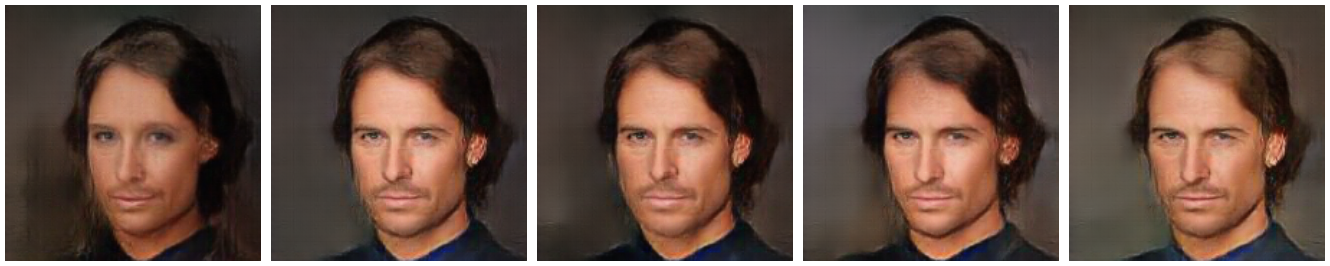} \\
        {\scriptsize \begin{tabularx}{0.32\textwidth}{CCCCC}\Female +beard & \Male -mkup & open-smile & \Male +bangs & \textcolor{gold}{hair}-glass \\ \end{tabularx}} & & {\scriptsize \begin{tabularx}{0.32\textwidth}{CCCCC}\Female +beard & \Male -mkup & open-smile & \Male +bangs & \textcolor{gold}{hair}-glass \\ \end{tabularx}} & & {\scriptsize \begin{tabularx}{0.32\textwidth}{CCCCC}\Female +beard & \Male -mkup & open-smile & \Male +bangs & \textcolor{gold}{hair}-glass \\ \end{tabularx}} \\ \vspace{-2mm}
        \includegraphics[width=0.32\textwidth]{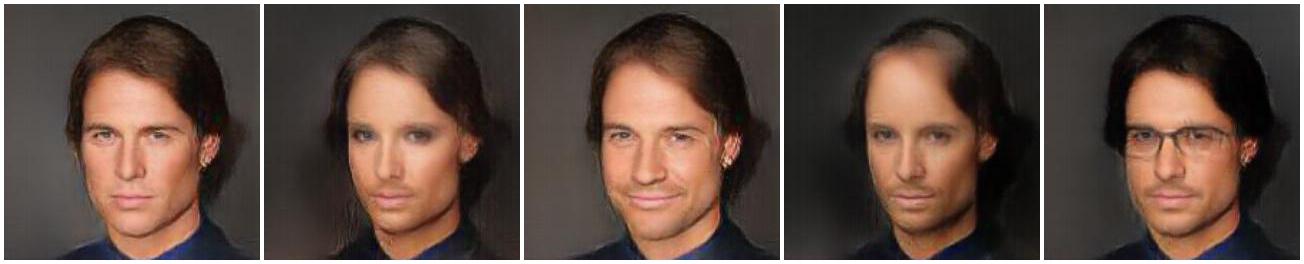} & & \includegraphics[width=0.32\textwidth]{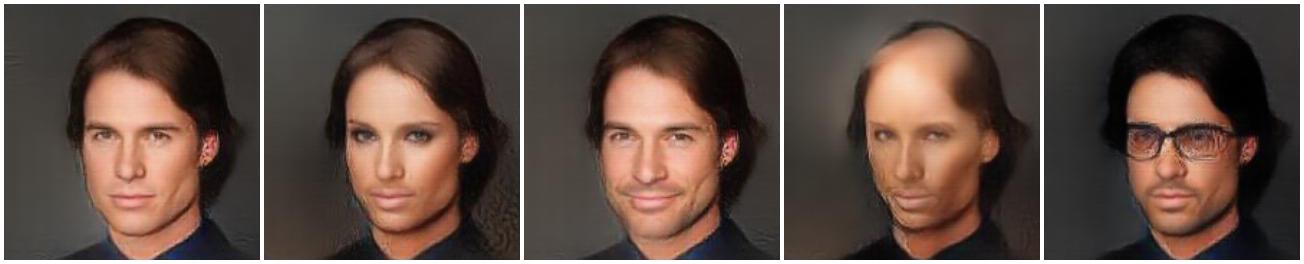} & & \includegraphics[width=0.32\textwidth]{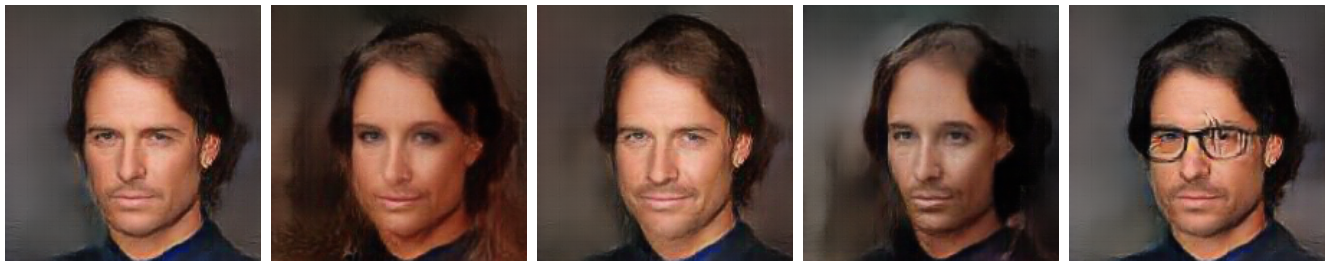} \\
        {\scriptsize \begin{tabularx}{0.32\textwidth}{CCCCC}\Male -beard & \Female +mkup & shut+smile & \Female +bald & hair+glass \\ \end{tabularx}} & & {\scriptsize \begin{tabularx}{0.32\textwidth}{CCCCC}\Male -beard & \Female +mkup & shut+smile & \Female +bald & hair+glass \\ \end{tabularx}} & & {\scriptsize \begin{tabularx}{0.32\textwidth}{CCCCC}\Male -beard & \Female +mkup & shut+smile & \Female +bald & hair+glass \\ \end{tabularx}} \\ \vspace{-2mm}
        \includegraphics[width=0.32\textwidth]{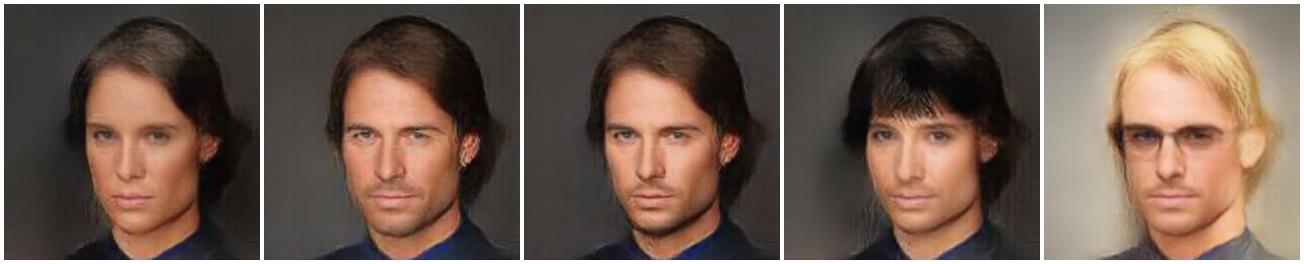} & & \includegraphics[width=0.32\textwidth]{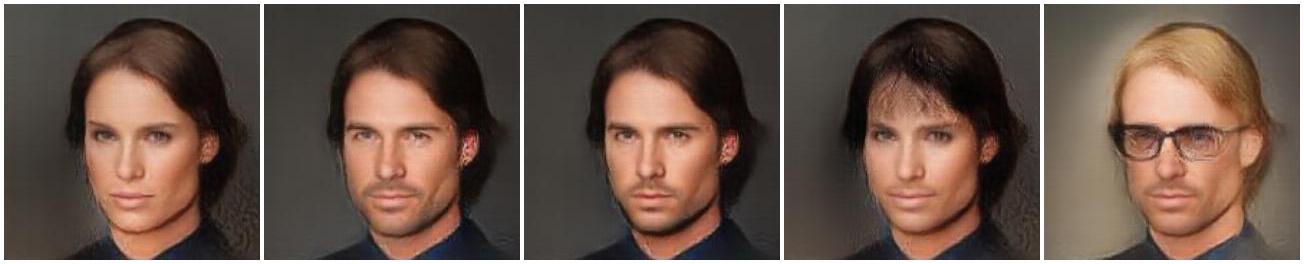} & & \includegraphics[width=0.32\textwidth]{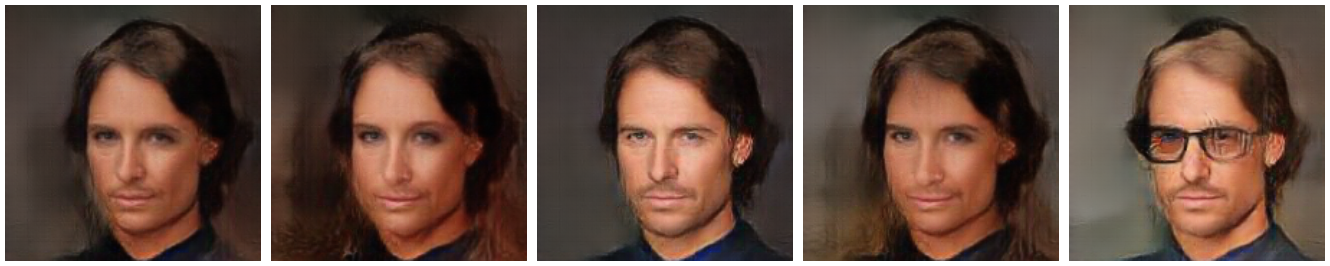} \\
        {\scriptsize \begin{tabularx}{0.32\textwidth}{CCCCC}\Female -beard & \Female -mkup & shut-smile & \Female +bangs &  \textcolor{gold}{hair}+glass \\ \end{tabularx}} & & {\scriptsize \begin{tabularx}{0.32\textwidth}{CCCCC}\Female -beard & \Female -mkup & shut-smile & \Female +bangs &  \textcolor{gold}{hair}+glass \\ \end{tabularx}} & & {\scriptsize \begin{tabularx}{0.32\textwidth}{CCCCC}\Female -beard & \Female -mkup & shut-smile & \Female +bangs &  \textcolor{gold}{hair}+glass \\ \end{tabularx}} \\ \vspace{-2mm}
    \end{tabularx}}
    \caption{Examples of image attribute manipulation using VAE backbone.}
    \label{fig:modifications}
    
\end{figure*}

\begin{table}[tb]
\centering
\caption{Accuracy and FID scores of image manipulation performed on the VAE backbone.}
\label{tab:image_manipulation}
\begin{tabular}{ccccc}
\toprule
Requested value  & \our{} & MSP & cFlow \\
\midrule
male x beard             & 0.80  & 0.79  & {\bf 0.85}  \\
female x beard           & {\bf 0.59}  & 0.33  & 0.31  \\
male x no-beard          & 0.88  & {\bf 0.92}  & 0.91  \\
female x no-beard        & 0.85  & 0.82  & {\bf 0.95}  \\
male x makeup            & {\bf 0.44}  & 0.43  & 0.29  \\
male x no-makeup         & 0.72  & 0.92  & {\bf 0.96}  \\
female x makeup          & 0.42  & 0.41  & {\bf 0.58}  \\
female x no-makeup       & 0.55  & 0.40  & {\bf 0.85}  \\
smile x open-mouth       & 0.97  & {\bf 0.99}  & 0.79  \\
no-smile x open-mouth    & 0.79  & {\bf 0.82}  & 0.77  \\
smile x calm-mouth       & 0.84  & {\bf 0.91}  & 0.72  \\
no-smile x calm-mouth    & 0.96  & 0.97  & {\bf 0.99}  \\
male x bald              & 0.26  & {\bf 0.41}  & 0.34  \\
male x bangs             & 0.58  & {\bf 0.74}  & 0.45  \\
female x bald            & 0.19  & 0.13  & {\bf 0.39}  \\
female x bangs           & 0.52  & 0.49  & {\bf 0.60}  \\
no-glasses x black-hair  & 0.92  & {\bf 0.93}  & 0.74  \\
no-glasses x golden-hair & {\bf 0.92}  & 0.91  & 0.81  \\
glasses x black-hair     & 0.76  & {\bf 0.90}  & 0.58  \\
glasses x golden-hair    & 0.75  & {\bf 0.85}  & 0.61  \\

\midrule
Average Acc  & 0.69 & {\bf 0.70} &  0.67 \\
Average FID  & {\bf 28.07} & 30.67 & 39.68 \\ 
\bottomrule
\end{tabular}
\end{table}

\paragraph{Image generation with VAE backbone}

In addition to manipulating the labeled attributes of existing images, \our{} generates new examples with desired attributes. To verify this property, we use the same VAE architecture as before trained on CelebA dataset. The baselines include cFlow and two previously introduced methods for multi-label conditional generation\footnote{For cVAE and $\Delta$-GAN we use images of the size $64 \times 64$ following their implementations.}: cVAE \cite{yan2016attribute2image} and $\Delta$-GAN \cite{gan2017triangle}. We exclude MSP from the comparison because it cannot generate new images, but only manipulate the attributes of existing ones (see supplementary materials for a detailed explanation).

\begin{figure}[t!]
    \centering
        \includegraphics[width=\linewidth]{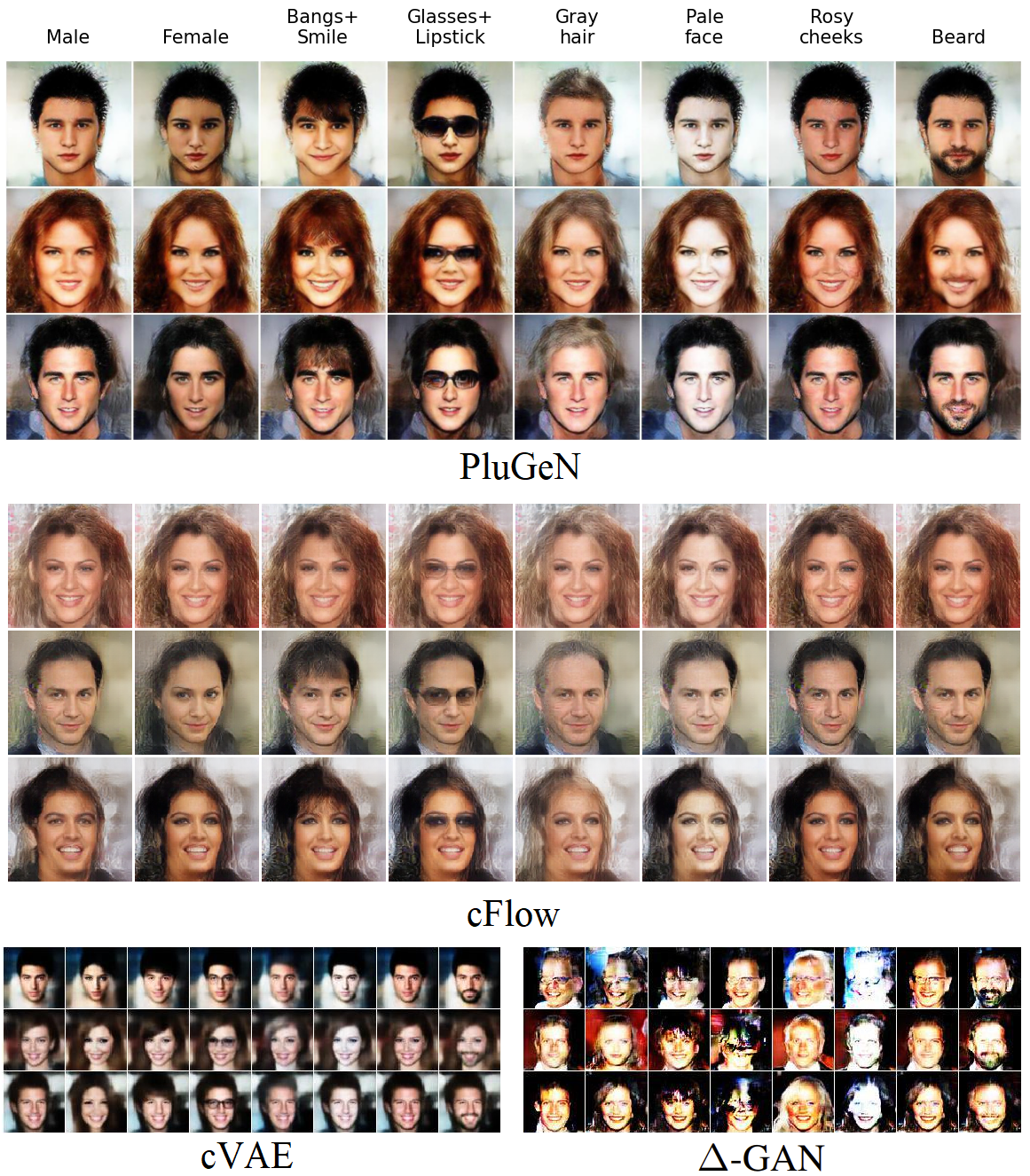}
    \caption{Examples of conditional generation using VAE backbone. Each row contains the same person (style variables) with modified attributes (label variables). }
    \label{fig:modified_label_generation}
\end{figure}

Figure~\ref{fig:modified_label_generation} presents sample results of image generation with the specific conditions. In each row, we fix the style variables $\mathbf{s}$ and vary the label variables $\mathbf{c}$ in each column, generating the same person but with different characteristics such as hair color, eyeglasses, etc. Although cVAE manages to modify the attributes, the quality of obtained samples is poor, while $\Delta$-GAN falls completely out of distribution. \our{} and cFlow generate images of similar quality, but only \our{} is able to correctly manipulate the labeled attributes. The lower quality of generated images is caused by the poor generation abilities of VAE backbone, which does not work well with high dimensional images (see supplementary materials for a discussion). For this reason, it is especially notable that \our{} can improve the generation performance of the backbone model in contrast to MSP.

\paragraph{Disentangling the attributes}

The attributes in the CelebA dataset are strongly correlated and at times even contradictory, e.g. attributes 'bald' and 'blond hair' cannot both be present at the same time.
In this challenging task, we aim to disentangle the attribute space as much as it is possible to allow for generating examples with arbitrary combinations of attributes. For this purpose, we sample the conditional variables $c_i$ independently, effectively ignoring the underlying correlations of attributes, and use them to generate images.
Since the attributes in the CelebA dataset are often imbalanced (e.g. only in 6.5\% of examples the person wears glasses), we calculate F1 and AUC scores for each attribute.

The quantitative analysis of the generated images presented in Table \ref{tab:disentanglement_metrics} confirms that \our{} outperforms the rest of the methods with respect to classification scores. The overall metrics are quite low for all approaches, which is due to the difficulty of disentanglement mentioned above, as well as the inaccuracy of the ResNet attribute classifier. Deep learning models often fail when the correlations in the training data are broken, e.g. the classifier might use the presence of a beard to predict gender, thus introducing noise in the evaluation \cite{Beery2018RecognitionIT}.

\begin{table}[tb]
\centering
\caption{Results of the independent conditional generation using VAE backbone.}
\label{tab:disentanglement_metrics}
\begin{tabular}{lrrrr}
    \toprule
                &  \our{} &  cFlow & $\Delta$-GAN & cVAE   \\
    \midrule
    F1  & \textbf{0.44}    & 0.29       & 0.39        & 0.39   \\
    AUC         & \textbf{0.76}     & 0.68       & 0.70        & 0.73   \\
    \bottomrule
    \end{tabular}
\end{table}

\paragraph{Chemical molecules modeling}

Finally, we present a practical use-case, in which we apply \our{} to generate chemical molecules with the requested properties.
As a backbone model, we use CharVAE~\cite{gomez2018automatic}, which is a type of recurrent network used for processing SMILES~\cite{weininger1988smiles}, a textual representation of molecules. It was trained on ZINC 250k database~\cite{sterling2015zinc} of commercially available chemical compounds. For every molecule, we model 3 physio-chemical continuous (not binary) labels: logP, SAS, TPSA, which values were calculated using RDKit package~\cite{landrum2006rdkit}. 
Additional explanations and more examples are given in the supplementary materials.

\begin{figure}[t!]
    \centering
    \includegraphics[width=0.49\textwidth]{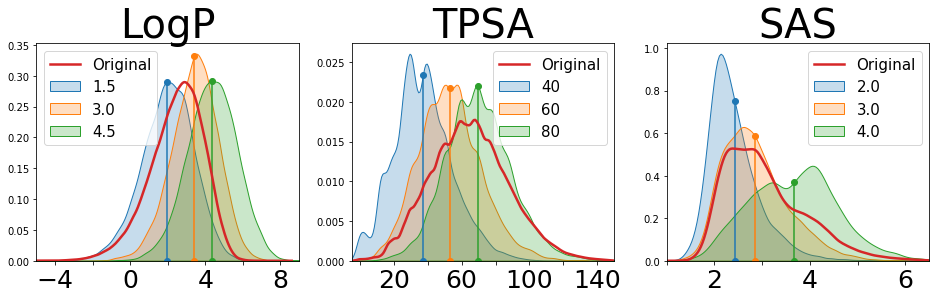}
    \caption{Distribution of attributes of generated molecules, together with distribution for the training dataset. Each color shows the value of a labeled attribute that was used for generation. \our{} is capable of moving the density of generated molecules' attributes towards the desired value. The average of every distribution is marked with a vertical line.}
    \label{fig:conditional_chemistry}
\end{figure}

First, we imitate a practical task of de novo design~\cite{olivecrona2017molecular,popova2018deep}, where we force the model to generate new compounds with desirable properties. For every attribute, we generate 25k molecules with 3 different values: for logP we set the label of generated molecules to: 1.5, 3.0, 4.5; for TPSA we set generated labels to: 40, 60, 80; for SAS we set them to: 2.0, 3.0, 4.0, which gives 9 scenarios in total. 
From density plots of labels of generated and original molecules presented in Figure~\ref{fig:conditional_chemistry}, we can see that \our{} changes the distribution of values of the attributes and moves it towards the desired value. A slight discrepancy between desired and generated values may follow from the fact that values of labeled attributes were sampled independently, which could make some combinations physically contradictory.

\begin{figure}[t!]
    \centering
    \subfloat[Molecules decoded from path]{
        \includegraphics[width=0.45\linewidth]{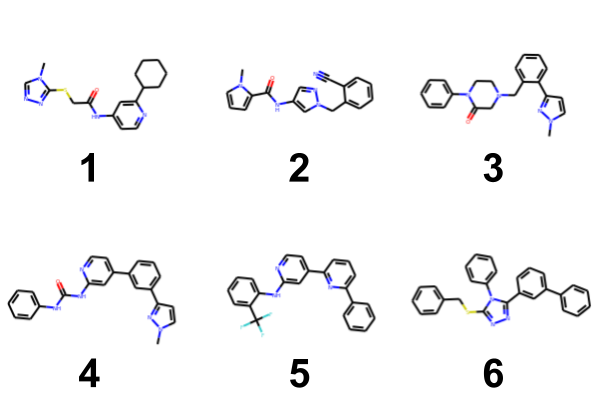}
    }
    \subfloat[LogP of presented molecules]{
        \includegraphics[width=0.45\linewidth]{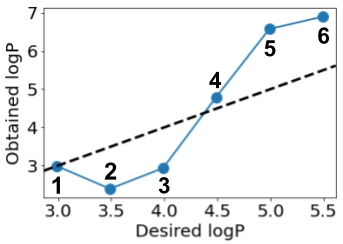}
    }\\
    \caption{Molecules obtained by the model during an optimization phase (left side), and their LogP (right side).}
    \label{fig:traversal_chemistry}
\end{figure}

Next, we consider the setting of lead optimization~\cite{jin2018learning,maziarka2020mol}, where selected compounds are improved to meet certain criteria. For this purpose, we encode a molecule into the latent representation of INF and force \our{} to gradually increase the value of logP by 3 and decode the resulting molecules. 
The obtained molecules together with their logP are shown in Figure~\ref{fig:traversal_chemistry}.
As can be seen, \our{} generates molecules that are structurally similar to the initial one, however with optimized desired attributes.

Obtained results show that \our{} is able to model the physio-chemical molecular features, which is a non-trivial task that could speed up a long and expensive process of designing new drugs.

\section{Conclusion} 

We proposed a novel approach for disentangling the latent space of pre-trained generative models, which works perfectly for generating new samples with desired conditions as well as for manipulating the attributes of existing examples. In contrast to previous works, we demonstrated that \our{} performs well across diverse domains, including chemical molecule modeling, and can be combined with various architectures, such as GANs and VAEs backbones.

\paragraph{Ethical considerations statement}
We did not identify ethical issues concerning our work, as we do not collect data, and we do not foresee malicious applications or societal harm. However, we believe that disentangling factors of variation can have a positive effect on reducing unjust correlations in the data. For example, even though CelebA dataset contains twice as many old men as old women, our method can generate an equal proportion of samples from those classes, thus avoiding amplifying bias in the data.

\section*{Acknowledgement}

We are thankful to the Reviewers and the Meta-Reviewer for their valuable comments and insightful feedback, which helped us to further improve the paper. The authors would also like to thank prof. Jacek Tabor for his invaluable contribution to our work, discussions, suggestions and criticism. 

The research of M. Wołczyk was supported by the Foundation for Polish Science co-financed by the European Union under the European Regional Development Fund in the POIR.04.04.00-00-14DE/18-00 project carried out within the Team-Net programme. The research of M. Proszewska was supported by the National Science Centre (Poland), grant no.  2018/31/B/ST6/00993. The research of Ł. Maziarka was supported by the National Science Centre (Poland), grant no. 2019/35/N/ST6/02125. The research of  M. Zieba was supported by the National Centre of Science (Poland) Grant No. 2020/37/B/ST6/03463. The research of R. Kurczab was supported by the Polish National Centre for Research and Development (Grant LIDER/37/0137/L-9/17/NCBR/2018). The research of M. \'Smieja was funded by the Priority Research Area DigiWorld under the program Excellence Initiative -– Research University at the Jagiellonian University in Kraków.

\bibliography{aaai22}

\clearpage

\section{Parametrization of \our{}}

\paragraph{Continuous attributes}

In this section, we show that \our{} can also be applied to the case of continuous labeled attributes. Without loss of generality, we assume that $y_i \in [-1,1]$. Analogically to the case of binary labels, we assume that the conditional distribution of label variable $C_i$ given $Y_i=y_i$ is parametrized by 
$$
p_{C_i|Y_i = y_i} = \N(y_i, \sigma),
$$
where $\sigma > 0$ is the user-defined parameter controlling smoothness.

Observe that by marginalizing out the label variable $Y_i$ over training set $(\mathbf{x}, \mathbf{y})$, we obtain:
$$
p_{C_i} = \frac{1}{|\mathcal{X}|} \sum_{(\mathbf{x},\mathbf{y})} \N(y_i, \sigma),
$$
which coincides with a 1-dimensional kernel density estimator (KDE). Although KDE does not work well in high dimensional spaces, it is a reliable estimate of the probability density function in the $1$-dimensional situation considered here. 

For high values of $\sigma$ there is a huge overlap between Gaussian components. This results in small penalties in terms of negative log-likelihood for incorrect assignments. From a practical perspective, we start the training process with high values of $\sigma$, which provides reasonable initialization of \our{}. Next, we gradually decrease $\sigma$ to match the correct assignments.

\paragraph{Modeling imbalanced binary labels}

\newcolumntype{C}{>{\centering\arraybackslash}X}
\begin{figure*}[t!]
   \centering
   \begin{tabularx}{\linewidth}{CC} {\bf \large \our{}} & {\bf \large StyleFlow} \end{tabularx}\\
   {\bf Age}\\
    \includegraphics[width=0.48\linewidth]{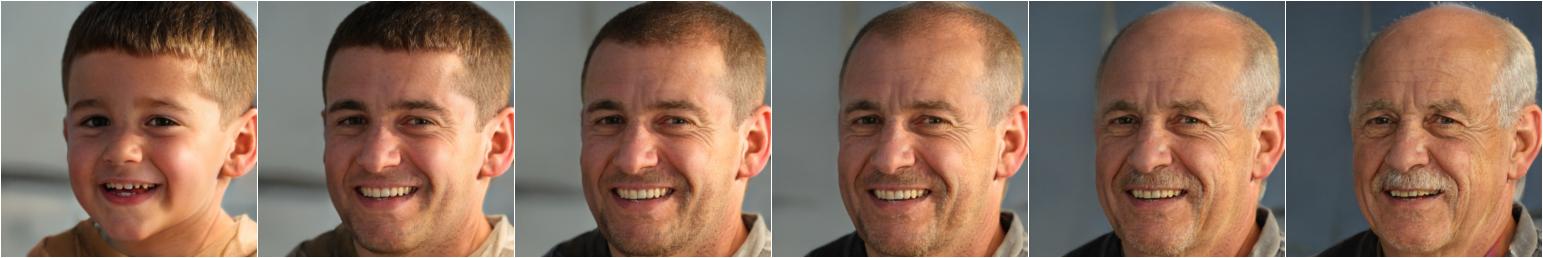} \quad \includegraphics[width=0.48\linewidth]{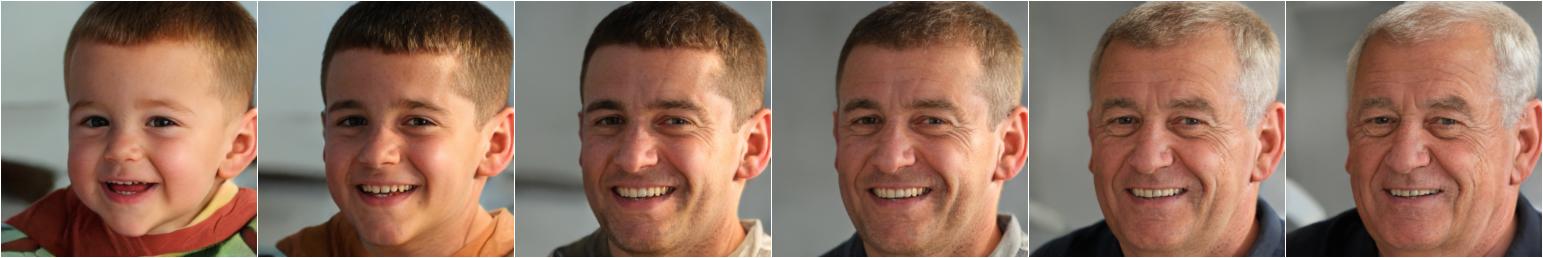}\\
    \vspace*{-0.2mm}
    \includegraphics[width=0.48\linewidth]{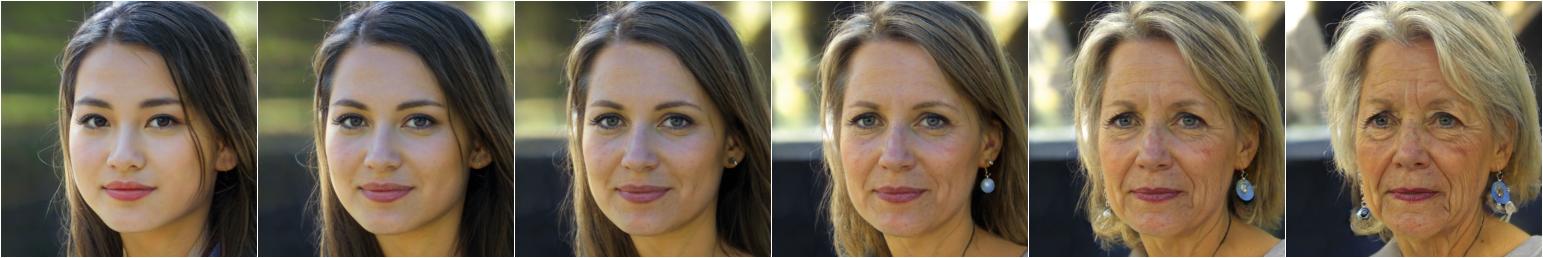} \quad \includegraphics[width=0.48\linewidth]{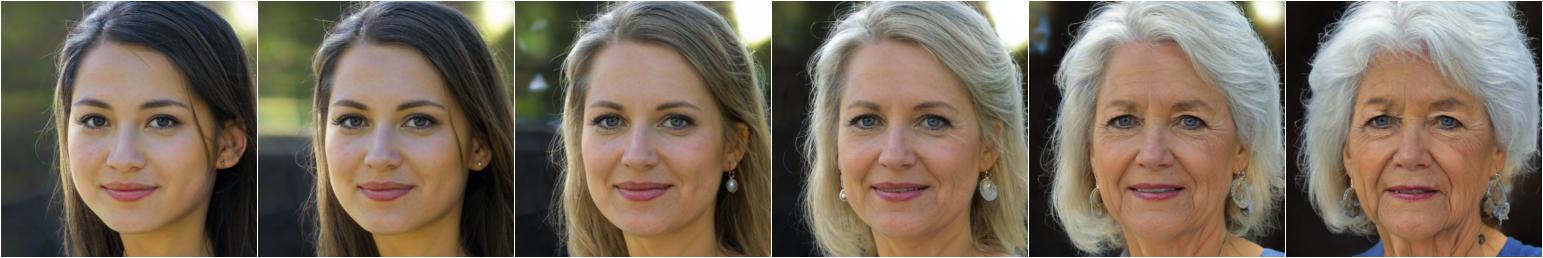}\\
    {\bf Baldness}\\
    \vspace*{-0.2mm}
    \includegraphics[width=0.48\linewidth]{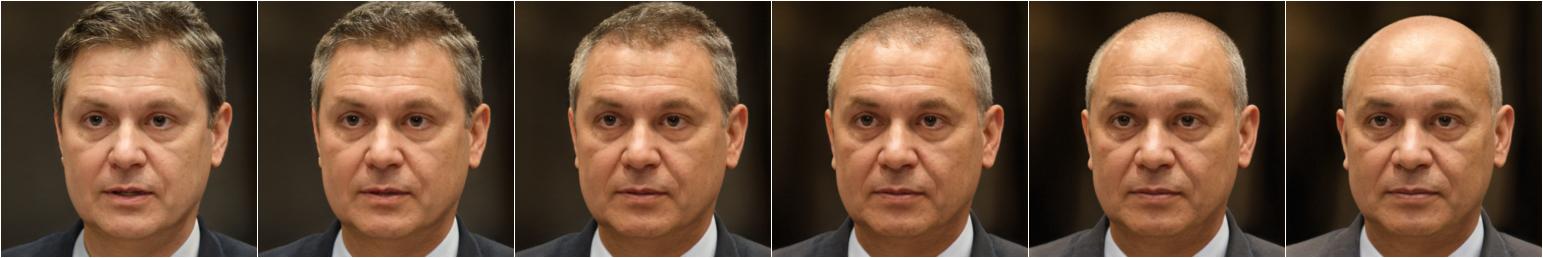} \quad \includegraphics[width=0.48\linewidth]{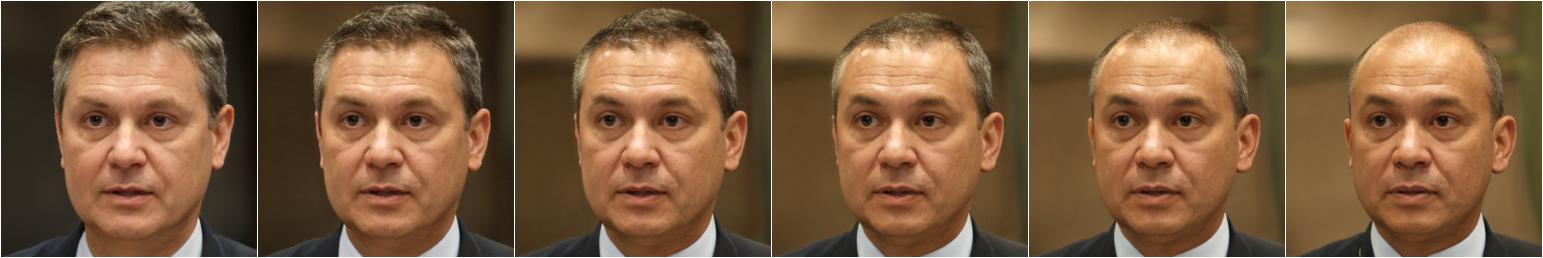}\\
    \vspace*{-0.2mm}
    \includegraphics[width=0.48\linewidth]{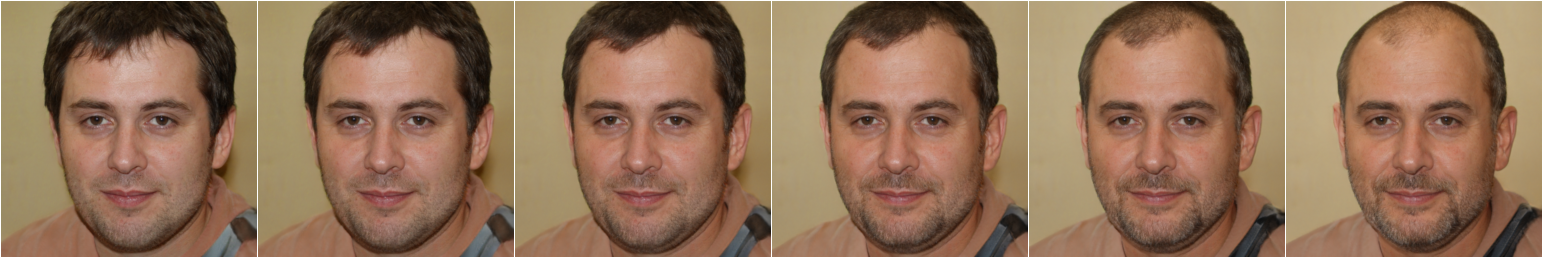} \quad \includegraphics[width=0.48\linewidth]{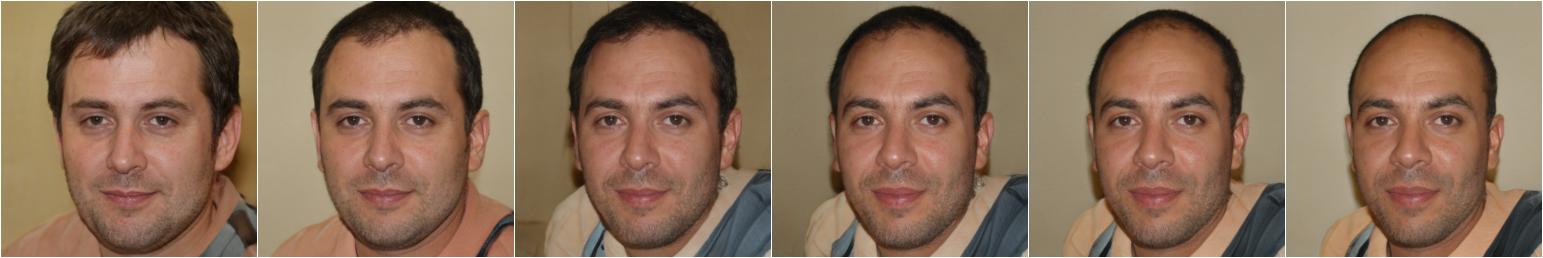}\\
    {\bf Beard}\\
    \vspace*{-0.2mm}
    \includegraphics[width=0.48\linewidth]{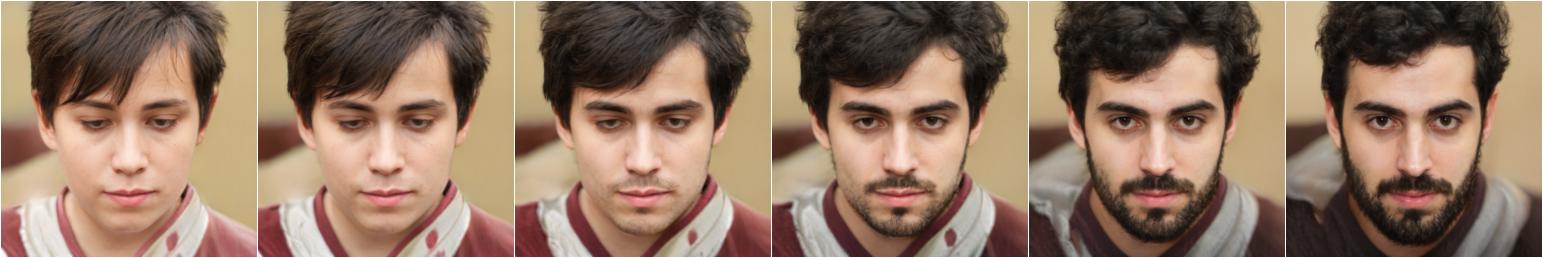} \quad \includegraphics[width=0.48\linewidth]{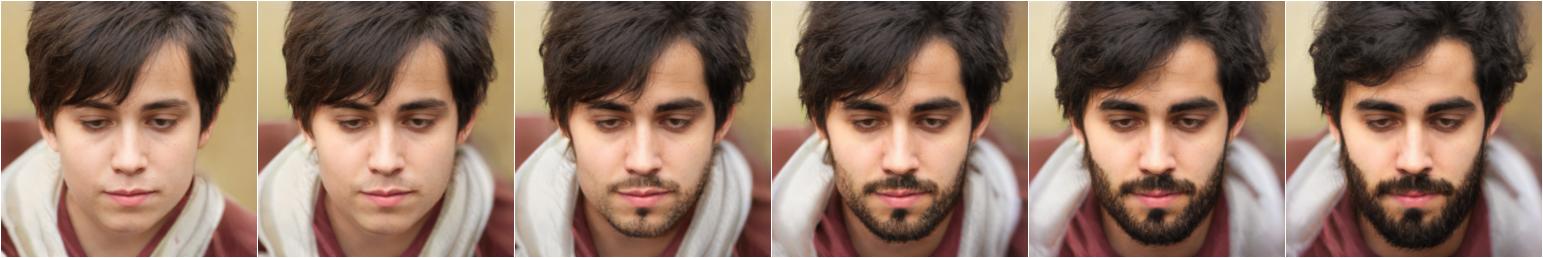}\\
    \vspace*{-0.2mm}
    \includegraphics[width=0.48\linewidth]{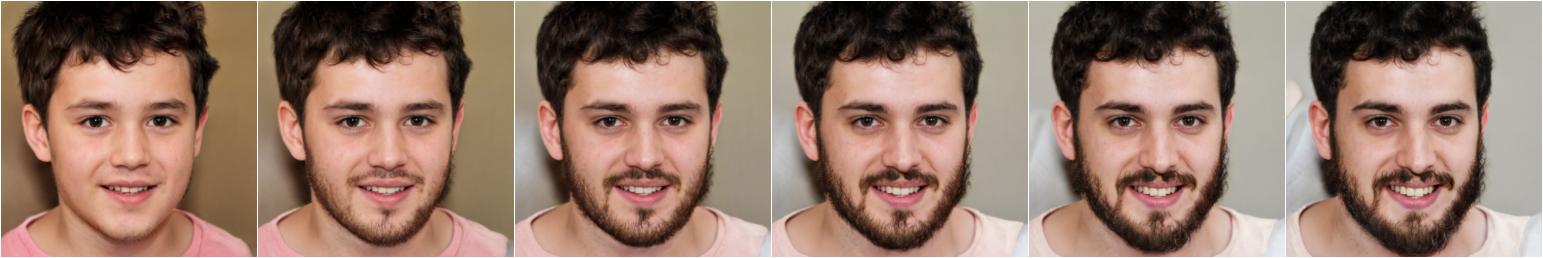} \quad \includegraphics[width=0.48\linewidth]{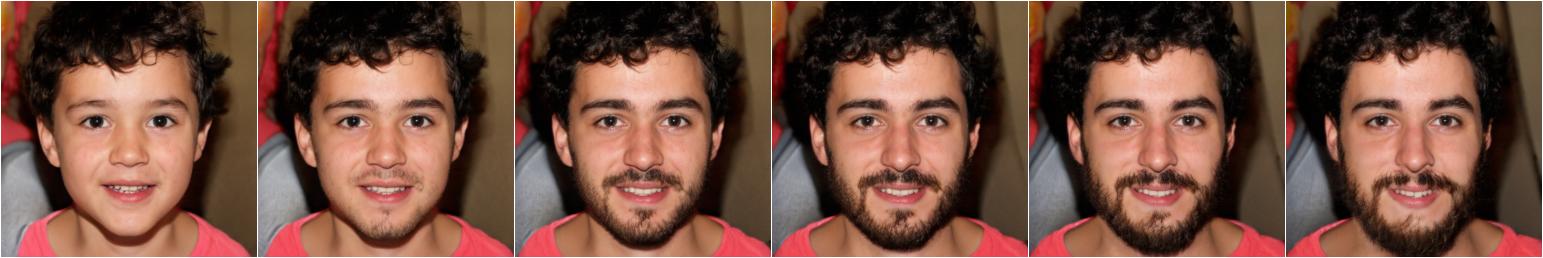} \\
    {\bf Yaw}\\
    \vspace*{-0.2mm}
    \includegraphics[width=0.48\linewidth]{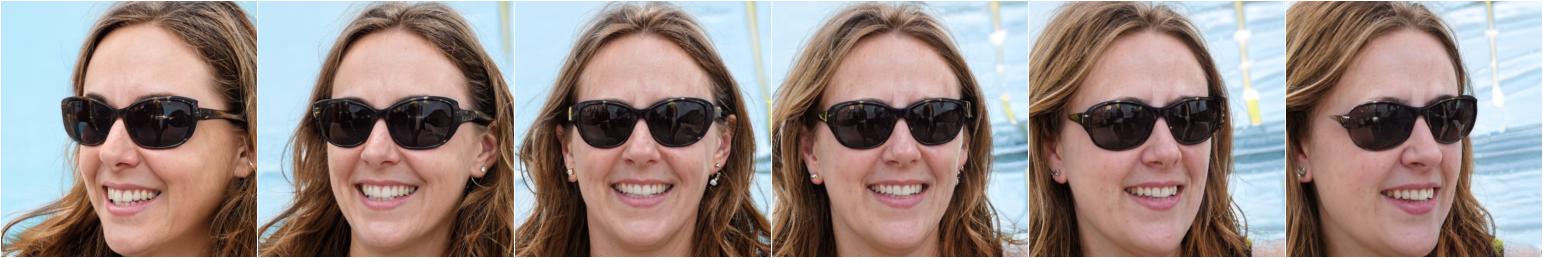} \quad \includegraphics[width=0.48\linewidth]{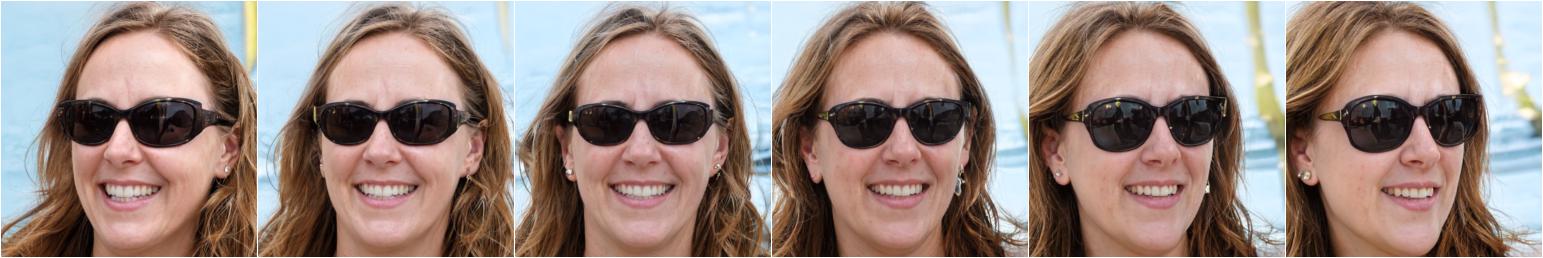}\\
    \vspace*{-0.2mm}
    \includegraphics[width=0.48\linewidth]{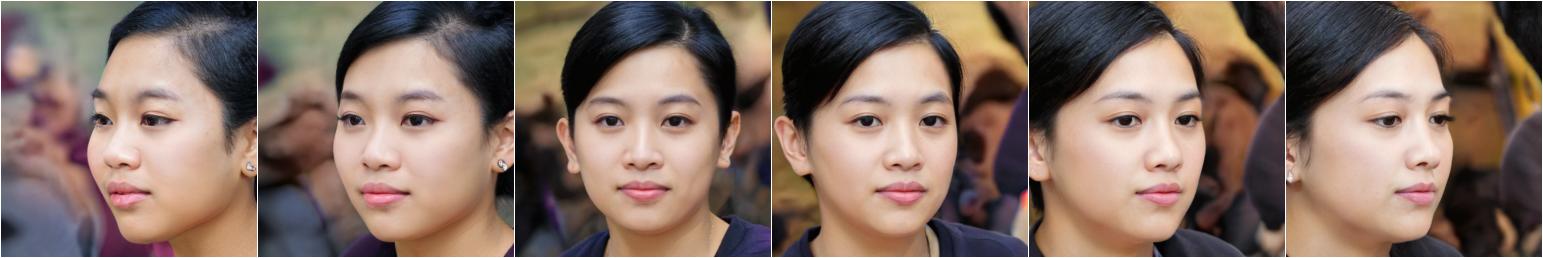} \quad \includegraphics[width=0.48\linewidth]{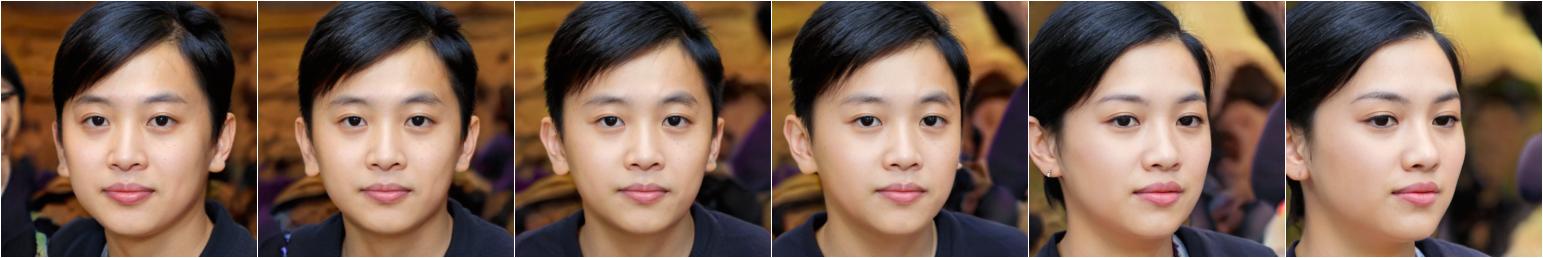}
    \caption{Gradual modification of attributes (age, baldness, beard, and yaw, respectively) performed by \our{} (left) and StyleFlow (right) using the StyleGAN backbone.}
    \label{fig:stylegan_interpolations}
  \end{figure*}

  \newcolumntype{C}{>{\centering\arraybackslash}X}
  \begin{figure*}[t!]
   \centering
   \begin{tabularx}{\linewidth}{CC} {\bf \large \our{}} & {\bf \large StyleFlow} \end{tabularx}
   \begin{tabularx}{0.48\linewidth}{CCCC} input & gender & \hspace*{-2mm} glasses & smile\end{tabularx} \quad \begin{tabularx}{0.48\linewidth}{CCCC} input & gender & \hspace*{-2mm} glasses & smile\end{tabularx}\\
    \includegraphics[width=0.48\linewidth]{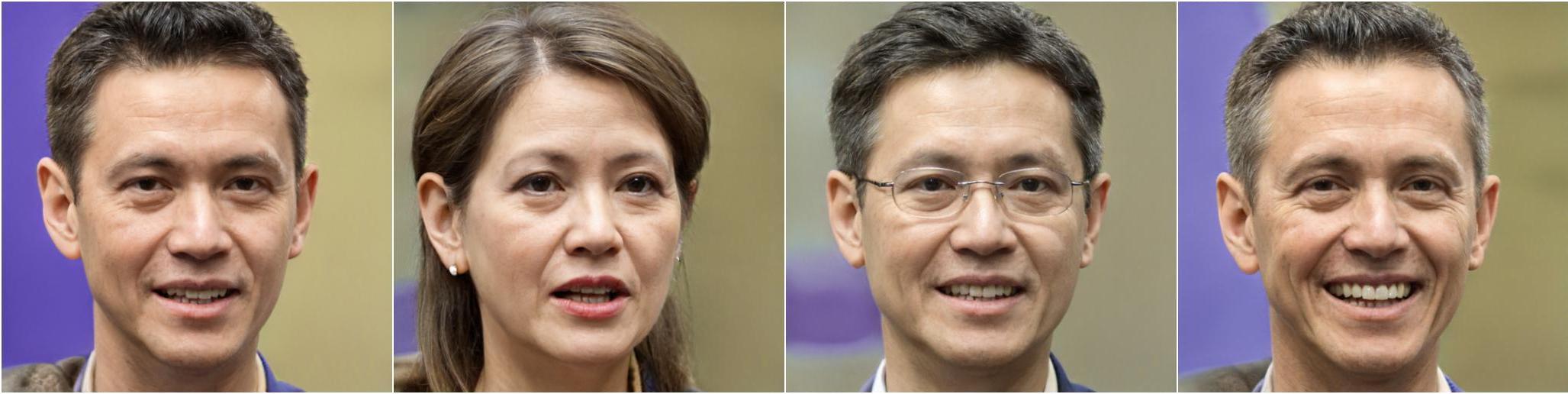} \quad \includegraphics[width=0.48\linewidth]{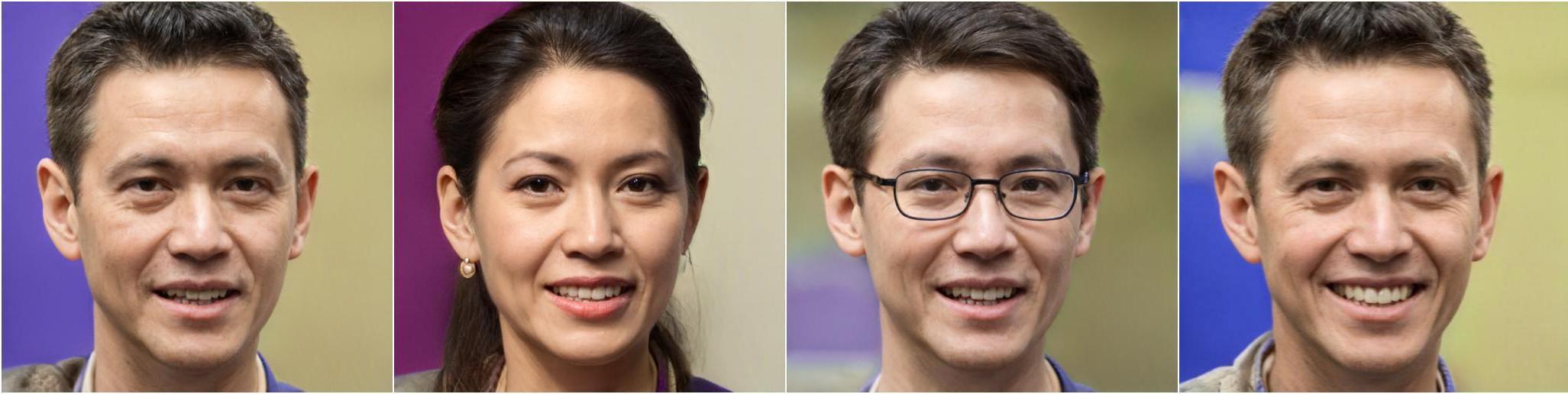}\\
    \vspace*{-0.4mm}
    \includegraphics[width=0.48\linewidth]{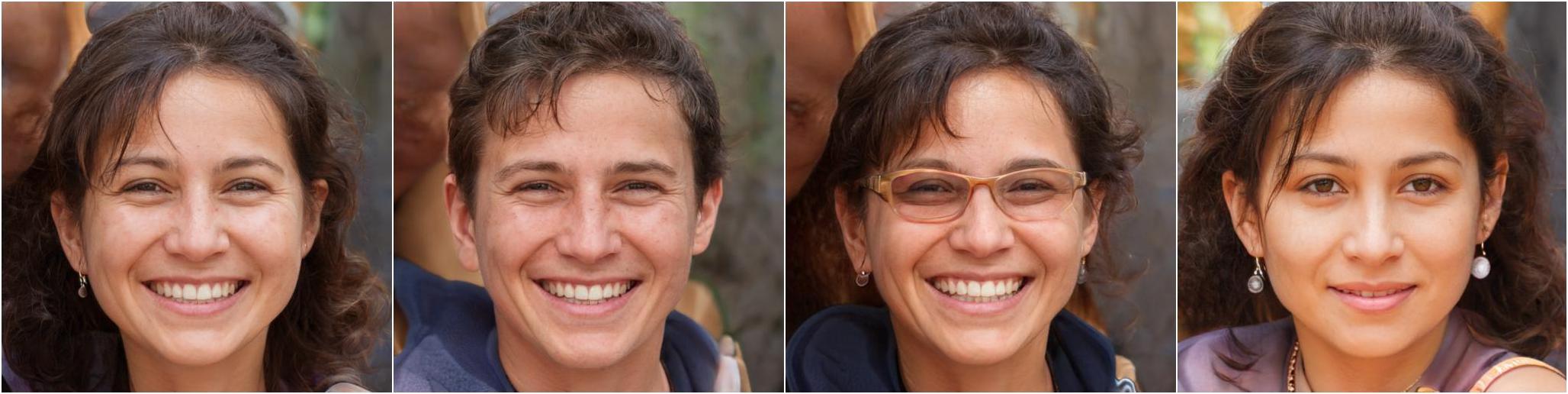} \quad \includegraphics[width=0.48\linewidth]{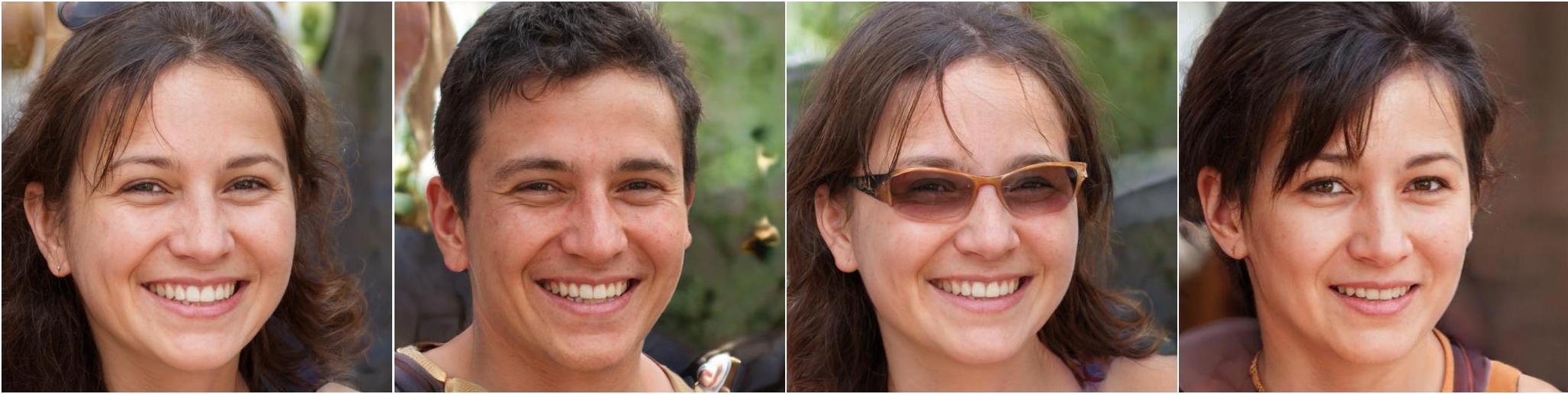}\\
    \vspace*{-0.4mm}
    \includegraphics[width=0.48\linewidth]{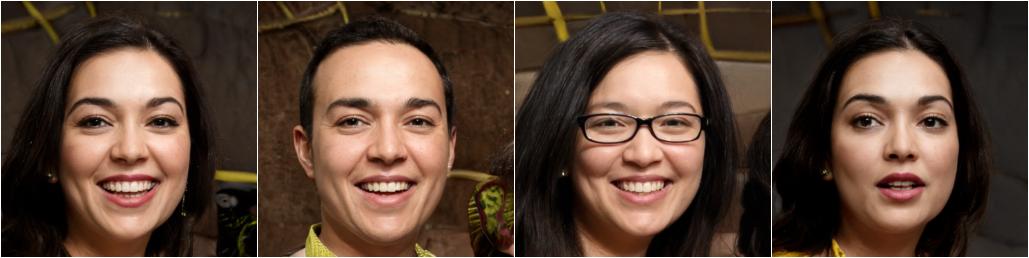} \quad \includegraphics[width=0.48\linewidth]{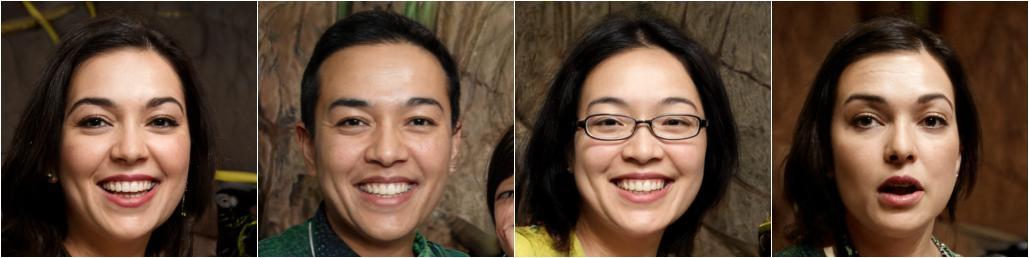}\\
    \vspace*{-0.4mm}
    \includegraphics[width=0.48\linewidth]{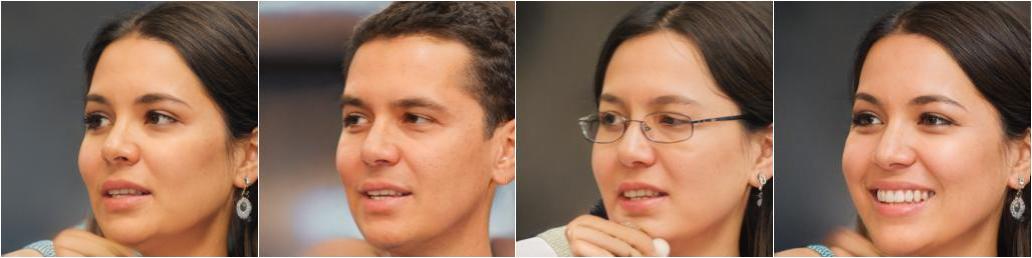} \quad \includegraphics[width=0.48\linewidth]{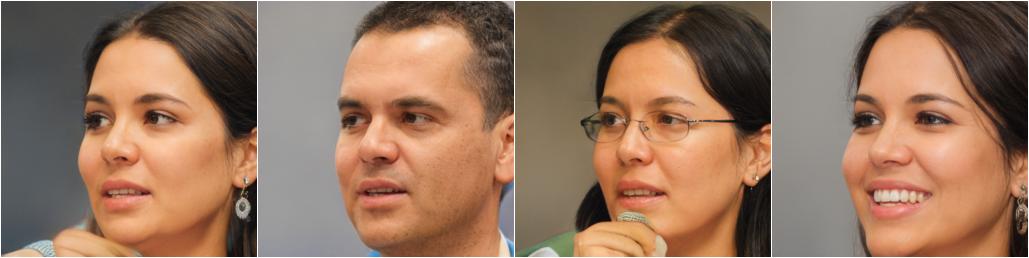}\\
    \caption{Attributes manipulation performed by \our{} (left) and StyleFlow (right) using the StyleGAN backbone.}
        \label{fig:stylegan_modifications}
 \end{figure*}

\definecolor{gold}{rgb}{0.85,.66,0}
\definecolor{grey}{rgb}{0.8, 0.8, 0.8}
\newcolumntype{C}{>{\centering\arraybackslash}X}
\begin{figure*}[ht]
    \subfloat{
    \setlength{\tabcolsep}{0pt}
    \begin{tabularx}{\linewidth}{cp{4pt}cp{4pt}c}
        {\begin{tabularx}{0.32\textwidth}{llC}
            \includegraphics[width=0.065\textwidth]{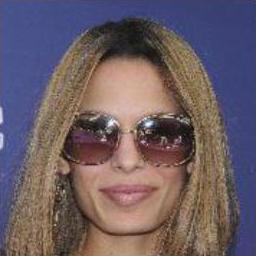} & \hspace{0.05cm} {\begin{turn}{90} \scriptsize Input image\end{turn}} & \vspace{-0.7cm} \hspace{-0.8cm} \our{}
        \end{tabularx}} & & MSP & & cFlow \\ \vspace{-2mm}
        \includegraphics[width=0.32\textwidth]{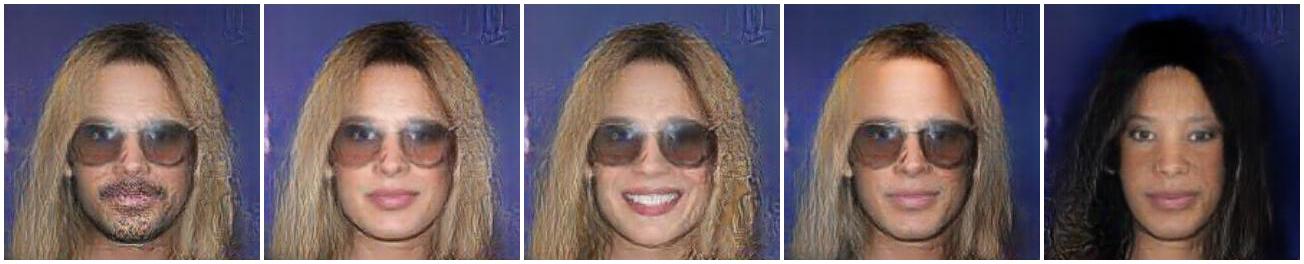} & & \includegraphics[width=0.32\textwidth]{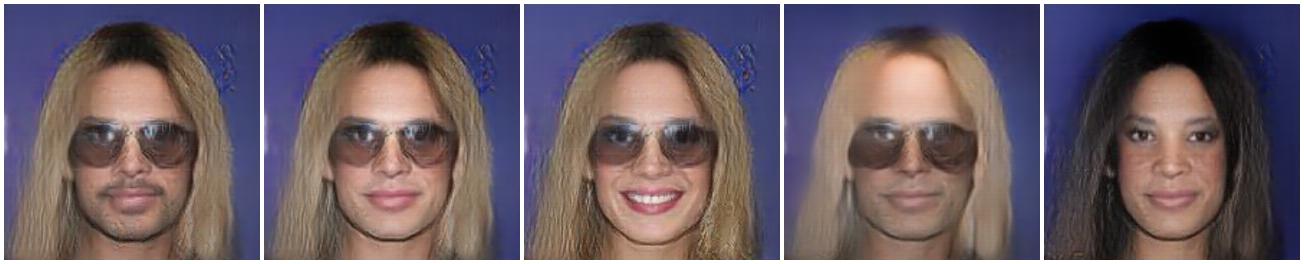} & & \includegraphics[width=0.32\textwidth]{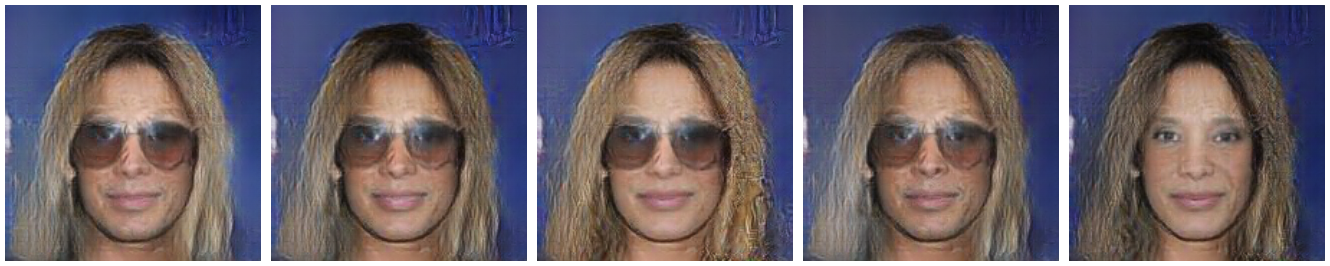} \\
        {\scriptsize \begin{tabularx}{0.32\textwidth}{CCCCC}\Male +beard & \Male +mkup & open+smile & \Male +bald & hair-glass \\ \end{tabularx}} & & {\scriptsize \begin{tabularx}{0.32\textwidth}{CCCCC}\Male +beard & \Male +mkup & open+smile & \Male +bald & hair-glass \\  \end{tabularx}} & & {\scriptsize \begin{tabularx}{0.32\textwidth}{CCCCC}\Male +beard & \Male +mkup & open+smile & \Male +bald & hair-glass \\ \end{tabularx}} \\ \vspace{-2mm}
        \includegraphics[width=0.32\textwidth]{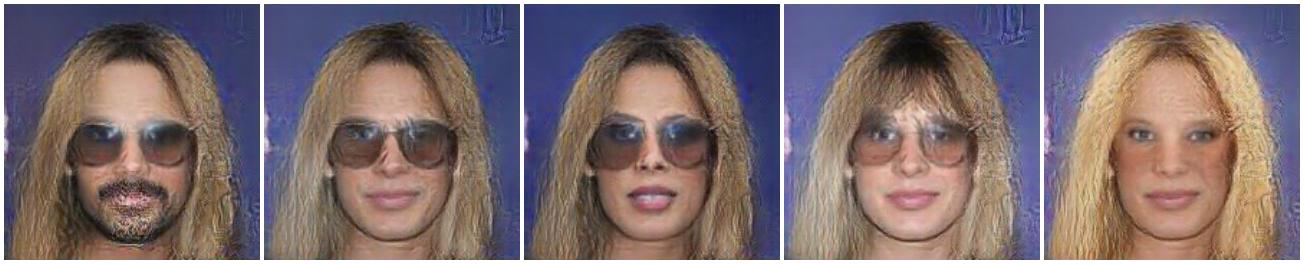} & & \includegraphics[width=0.32\textwidth]{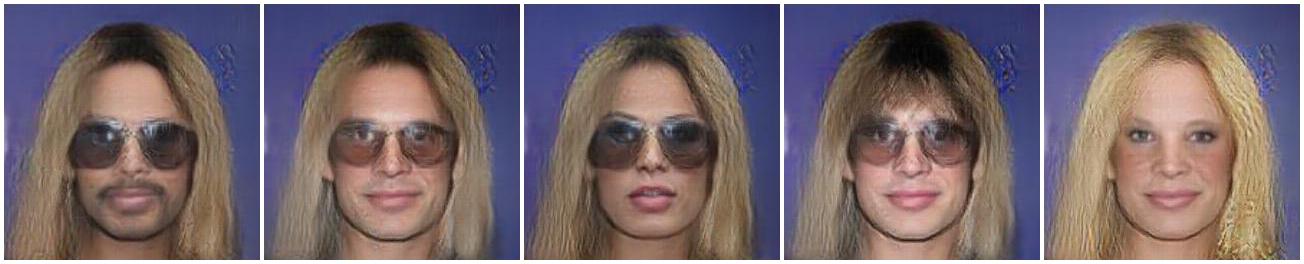} & & \includegraphics[width=0.32\textwidth]{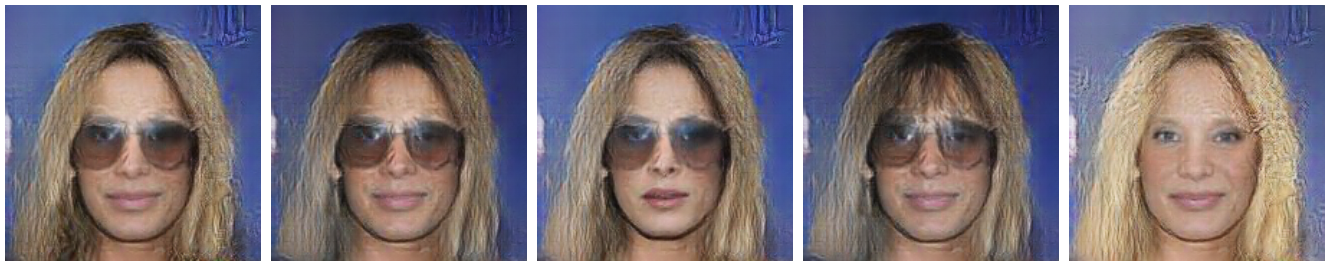} \\
        {\scriptsize \begin{tabularx}{0.32\textwidth}{CCCCC}\Female +beard & \Male -mkup & open-smile & \Male +bangs & \textcolor{gold}{hair}-glass \\ \end{tabularx}} & & {\scriptsize \begin{tabularx}{0.32\textwidth}{CCCCC}\Female +beard & \Male -mkup & open-smile & \Male +bangs & \textcolor{gold}{hair}-glass \\ \end{tabularx}} & & {\scriptsize \begin{tabularx}{0.32\textwidth}{CCCCC}\Female +beard & \Male -mkup & open-smile & \Male +bangs & \textcolor{gold}{hair}-glass \\ \end{tabularx}} \\ \vspace{-2mm}
        \includegraphics[width=0.32\textwidth]{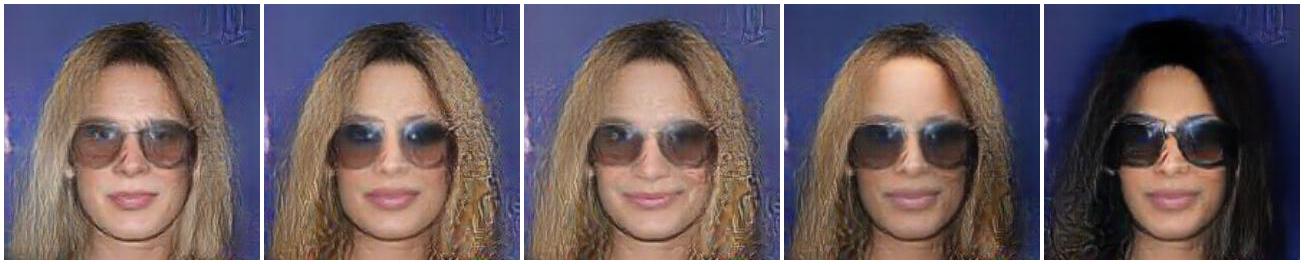} & & \includegraphics[width=0.32\textwidth]{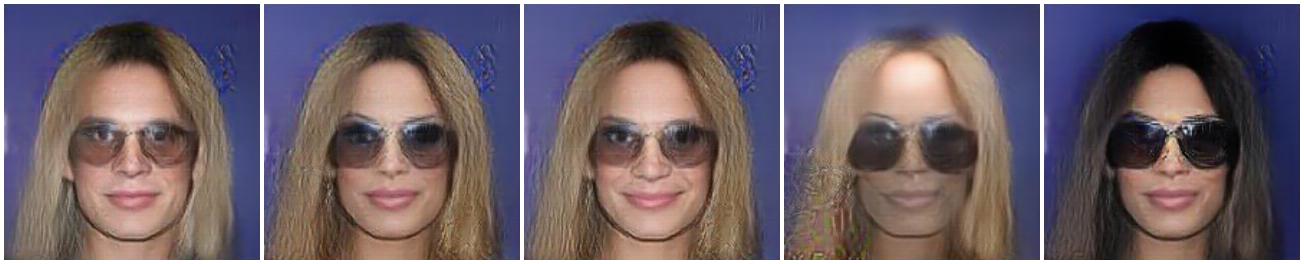} & & \includegraphics[width=0.32\textwidth]{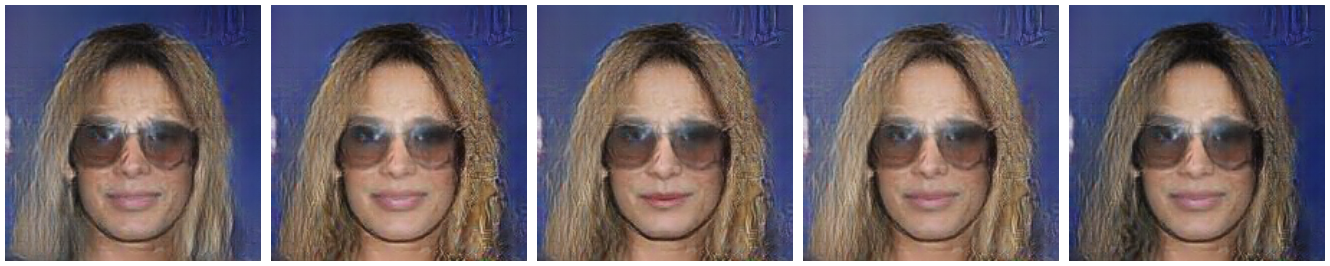} \\
        {\scriptsize \begin{tabularx}{0.32\textwidth}{CCCCC}\Male -beard & \Female +mkup & shut+smile & \Female +bald & hair+glass \\ \end{tabularx}} & & {\scriptsize \begin{tabularx}{0.32\textwidth}{CCCCC}\Male -beard & \Female +mkup & shut+smile & \Female +bald & hair+glass \\ \end{tabularx}} & & {\scriptsize \begin{tabularx}{0.32\textwidth}{CCCCC}\Male -beard & \Female +mkup & shut+smile & \Female +bald & hair+glass \\ \end{tabularx}} \\ \vspace{-2mm}
        \includegraphics[width=0.32\textwidth]{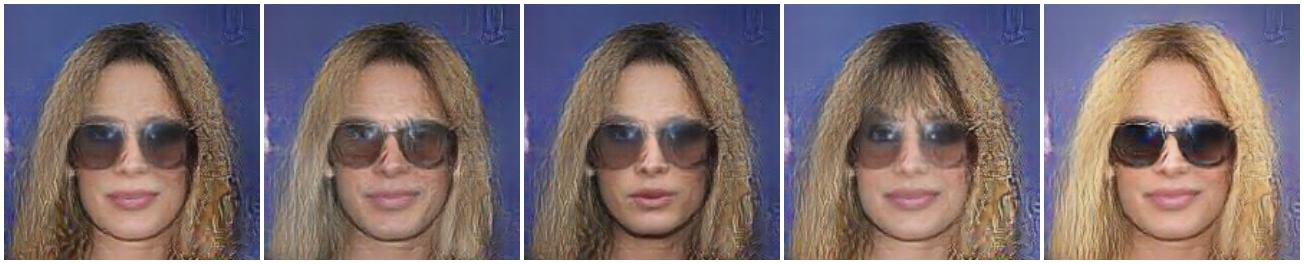} & & \includegraphics[width=0.32\textwidth]{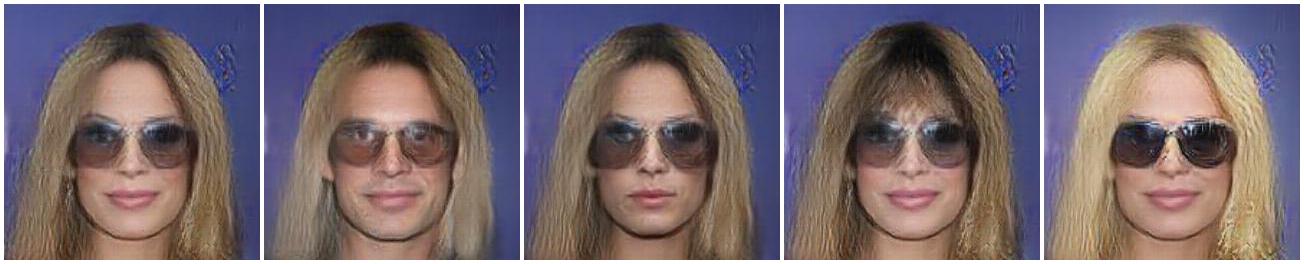} & & \includegraphics[width=0.32\textwidth]{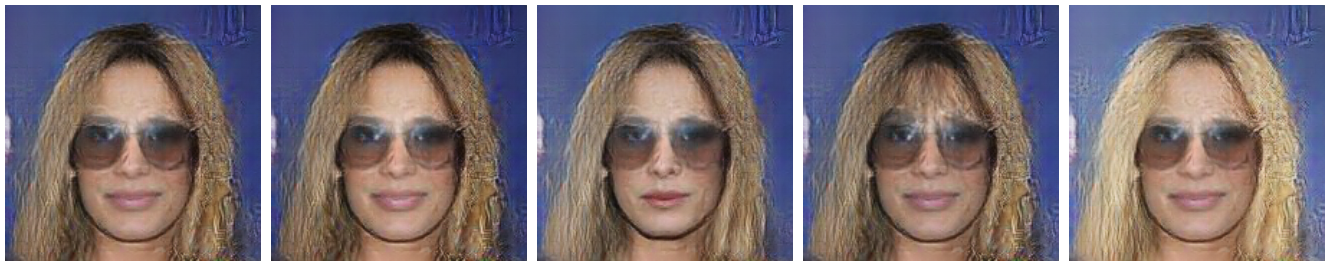} \\
        {\scriptsize \begin{tabularx}{0.32\textwidth}{CCCCC}\Female -beard & \Female -mkup & shut-smile & \Female +bangs &  \textcolor{gold}{hair}+glass \\ \end{tabularx}} & & {\scriptsize \begin{tabularx}{0.32\textwidth}{CCCCC}\Female -beard & \Female -mkup & shut-smile & \Female +bangs &  \textcolor{gold}{hair}+glass \\ \end{tabularx}} & & {\scriptsize \begin{tabularx}{0.32\textwidth}{CCCCC}\Female -beard & \Female -mkup & shut-smile & \Female +bangs &  \textcolor{gold}{hair}+glass \\ \end{tabularx}} \\ \vspace{-2mm}
    \end{tabularx}}
    
    \subfloat{
    \setlength{\tabcolsep}{0pt}
    \begin{tabularx}{\linewidth}{cp{4pt}cp{4pt}c}
        {\begin{tabularx}{0.32\textwidth}{llC}
            \includegraphics[width=0.065\textwidth]{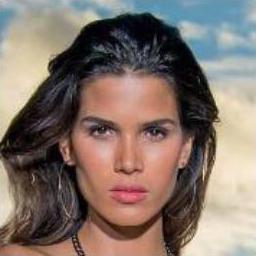} & \hspace{0.05cm} {\begin{turn}{90} \scriptsize Input image\end{turn}} & \vspace{-0.7cm} \hspace{-0.8cm} \our{}
        \end{tabularx}} & & MSP & & cFlow \\ \vspace{-2mm}
        \includegraphics[width=0.32\textwidth]{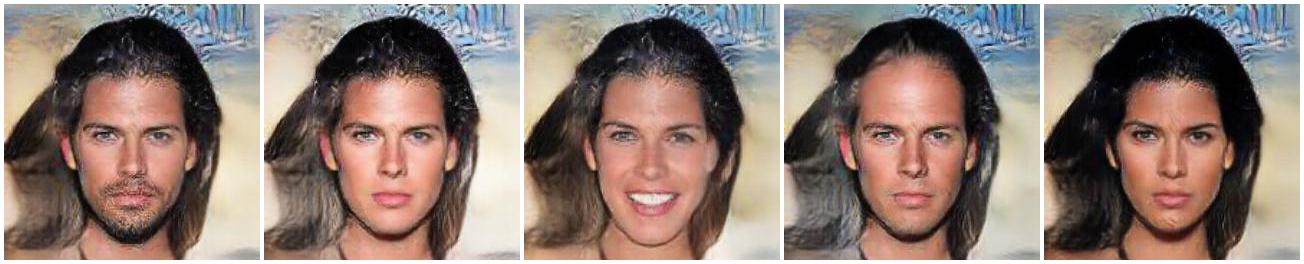} & & \includegraphics[width=0.32\textwidth]{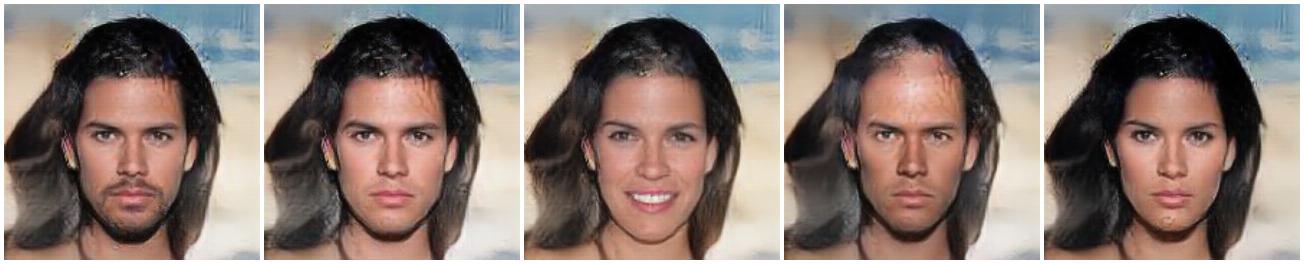} & & \includegraphics[width=0.32\textwidth]{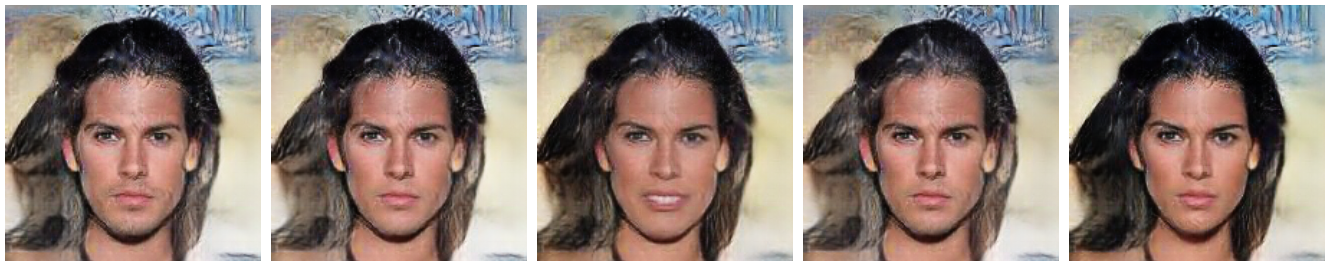} \\
        {\scriptsize \begin{tabularx}{0.32\textwidth}{CCCCC}\Male +beard & \Male +mkup & open+smile & \Male +bald & hair-glass \\ \end{tabularx}} & & {\scriptsize \begin{tabularx}{0.32\textwidth}{CCCCC}\Male +beard & \Male +mkup & open+smile & \Male +bald & hair-glass \\  \end{tabularx}} & & {\scriptsize \begin{tabularx}{0.32\textwidth}{CCCCC}\Male +beard & \Male +mkup & open+smile & \Male +bald & hair-glass \\ \end{tabularx}} \\ \vspace{-2mm}
        \includegraphics[width=0.32\textwidth]{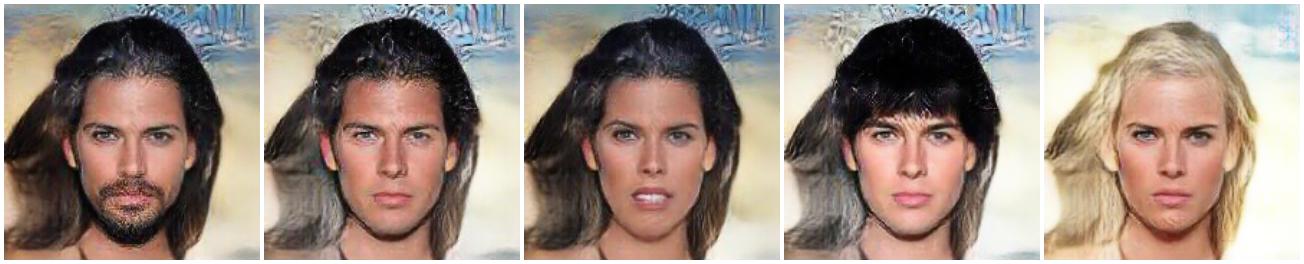} & & \includegraphics[width=0.32\textwidth]{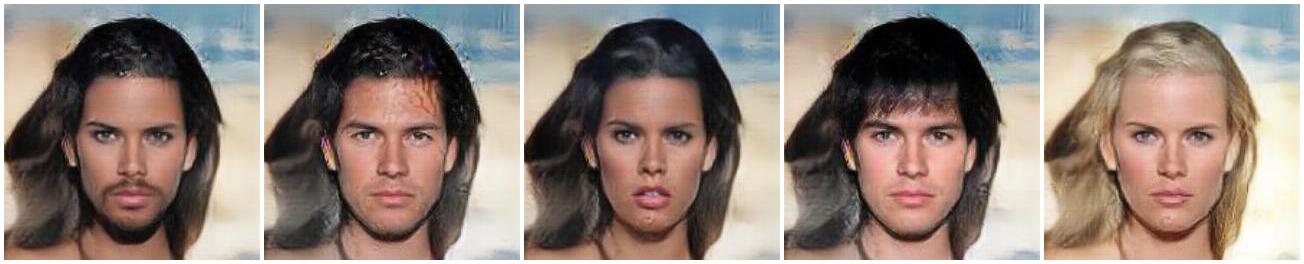} & & \includegraphics[width=0.32\textwidth]{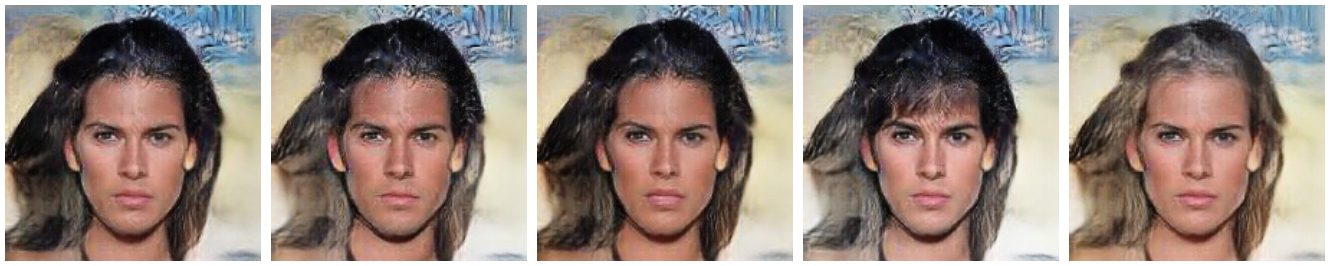} \\
        {\scriptsize \begin{tabularx}{0.32\textwidth}{CCCCC}\Female +beard & \Male -mkup & open-smile & \Male +bangs & \textcolor{gold}{hair}-glass \\ \end{tabularx}} & & {\scriptsize \begin{tabularx}{0.32\textwidth}{CCCCC}\Female +beard & \Male -mkup & open-smile & \Male +bangs & \textcolor{gold}{hair}-glass \\ \end{tabularx}} & & {\scriptsize \begin{tabularx}{0.32\textwidth}{CCCCC}\Female +beard & \Male -mkup & open-smile & \Male +bangs & \textcolor{gold}{hair}-glass \\ \end{tabularx}} \\ \vspace{-2mm}
        \includegraphics[width=0.32\textwidth]{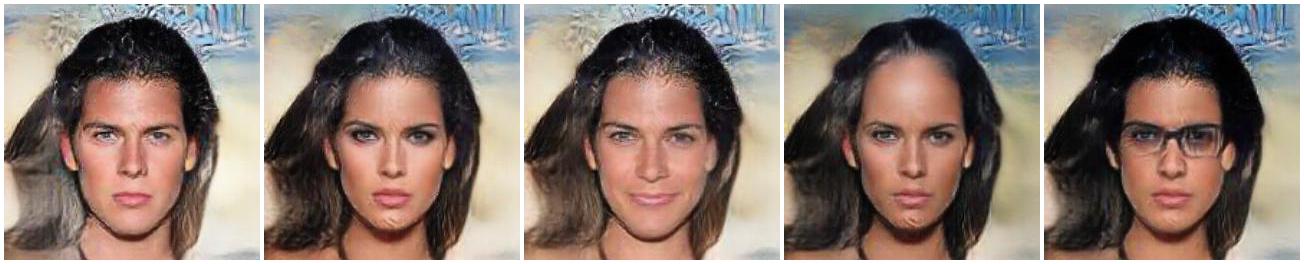} & & \includegraphics[width=0.32\textwidth]{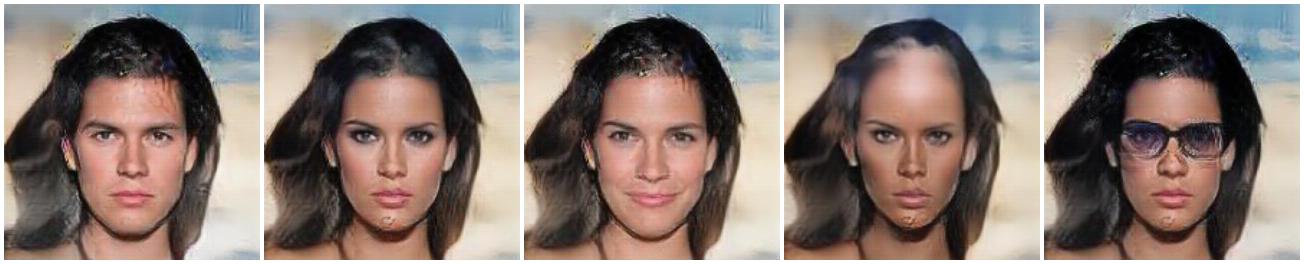} & & \includegraphics[width=0.32\textwidth]{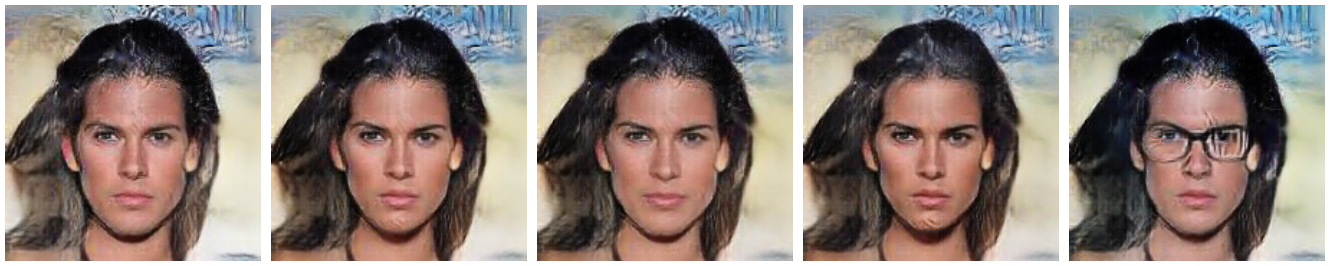} \\
        {\scriptsize \begin{tabularx}{0.32\textwidth}{CCCCC}\Male -beard & \Female +mkup & shut+smile & \Female +bald & hair+glass \\ \end{tabularx}} & & {\scriptsize \begin{tabularx}{0.32\textwidth}{CCCCC}\Male -beard & \Female +mkup & shut+smile & \Female +bald & hair+glass \\ \end{tabularx}} & & {\scriptsize \begin{tabularx}{0.32\textwidth}{CCCCC}\Male -beard & \Female +mkup & shut+smile & \Female +bald & hair+glass \\ \end{tabularx}} \\ \vspace{-2mm}
        \includegraphics[width=0.32\textwidth]{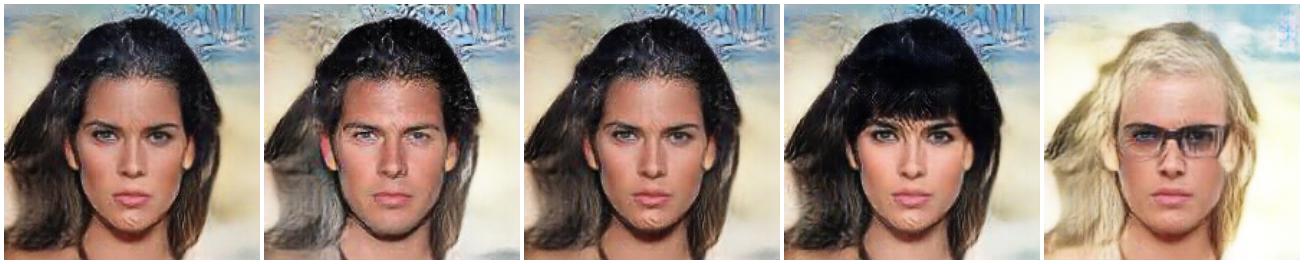} & & \includegraphics[width=0.32\textwidth]{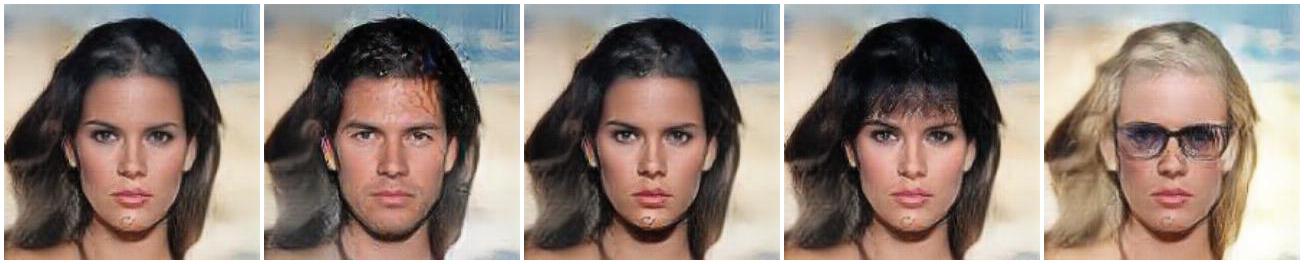} & & \includegraphics[width=0.32\textwidth]{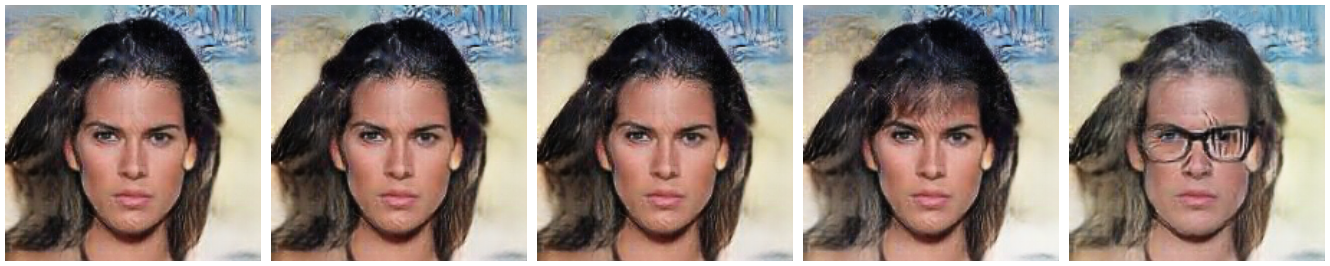} \\
        {\scriptsize \begin{tabularx}{0.32\textwidth}{CCCCC}\Female -beard & \Female -mkup & shut-smile & \Female +bangs &  \textcolor{gold}{hair}+glass \\ \end{tabularx}} & & {\scriptsize \begin{tabularx}{0.32\textwidth}{CCCCC}\Female -beard & \Female -mkup & shut-smile & \Female +bangs &  \textcolor{gold}{hair}+glass \\ \end{tabularx}} & & {\scriptsize \begin{tabularx}{0.32\textwidth}{CCCCC}\Female -beard & \Female -mkup & shut-smile & \Female +bangs &  \textcolor{gold}{hair}+glass \\ \end{tabularx}} \\ \vspace{-2mm}
    \end{tabularx}}
    \caption{Examples of image attribute manipulation using VAE backbone.}
    \label{fig:modifications2}
    
\end{figure*}

In many cases, the class labels are imbalanced, which means that the number of examples from one class significantly exceeds the other class (e.g., only $6.5\%$ examples in CelebA dataset have the 'glasses' label). To deal with imbalanced data, we scale the variance of Gaussian density modeling the conditional distribution $p_{C_i|Y_i=y_i}$.

We consider the conditional density of $i$-th attribute represented by:
\begin{equation}
p_{C_i|Y_i=y_i} = \N(m_0, \sigma_0)^{(1-y_i)} \cdot \N(m_1, \sigma_1)^{y_i},
\end{equation}
where $m_0=-1$ and $m_1=1$. We assume that $p_0,p_1$ are the fractions of examples with class 0 and 1, respectively. To deal with imbalanced classes we put 
$$
\sigma_i = \sigma \sqrt{2 p_i}, 
$$
where $\sigma > 0$ is a fixed parameter. For a majority class, standard deviation becomes higher, which introduces a lower penalty in the case of negative log-likelihood loss. The minority class has a higher penalty because we need to stop the mixture from collapsing  into a single component.

Let us calculate the log-likelihood of our conditional prior density $p_{C_i|Y_i=y_i}$ using the parametrization $\sigma_i = \sigma \sqrt{2 p_i}$. We have
\begin{multline} \label{eq:mix}
-\log p_{C_i|Y_i=y_i}(c) = y_i \cdot \lambda_1 \frac{(c - m_1)^2}{2\sigma^2} + \\
(1 - y_i) \cdot \lambda_0 \frac{(c-m_0)^2}{2 \sigma^2} + const,
\end{multline}
where $\lambda_i = \frac{1}{2 p_i}$ is an extra weighting factor.

We observe that, for our selection of $\sigma_i$, the expected value of the weighting factors with respect to labeling variable $y$ equals 1. In consequence,
\begin{multline*}
\mathbb{E}_{y}[-\log p_{C_i|Y_i=y_i}(c)] = \\
p_0 \frac{(c - m_1)^2}{2\sigma^2} + 
p_1 \frac{(c-m_0)^2}{2 \sigma^2} + const,
\end{multline*}
which is a typical log-likelihood of Gaussian distribution assuming class proportion $p_i$.

\paragraph{Reducing $\sigma$ in a training of \our{}}

Here, we describe the schedule for parameter $\sigma$ used for modeling conditional distribution $\N(m, \sigma)$. We want to ensure the flexibility of the INF at the beginning of the training, but we also need the attribute values to be strictly separated. In order to achieve both of these conditions, we impose a schedule on the standard deviation $\sigma$. Starting with high $\sigma$ we allow for great flexibility of our model, and then we get class separation by reducing the value of $\sigma$. Namely, we use the following schedule for the standard deviation $\sigma$ of the class normal distributions:
$$
\sigma(t) = \sigma_0 \cdot \gamma^{t}, 
$$
where $t$ is the index of the current epoch and $\sigma_0, \gamma$ are hyperparameters setting, respectively, the starting point and the speed of value decay. The selection process of $\sigma_0$ and $\gamma$ is described in the following sections.

\subsection{Semi-supervised setting}

It is worth noticing that \our{} can be trained in a semi-supervised setting, where only partial information about labeled attributes is available. Namely, for every latent representation $\mathbf{z}$ we can use an arbitrary condition imposed on $\mathbf{Y}$. If the value of $i$-th label is unknown, then we use the marginal density:
$$
p_{C_i} = p_0 \N(m_0, \sigma_0) + p_1 \N(m_1, \sigma_1)
$$
instead of $p_{C_i|Y_i}$ in the loss function (3). Here $p_0,p_1$ are the fraction of negatively and positively labeled examples in $\X$. Further investigations about this setting are left for future work.

\section{Details of image experiments}

All experiments were run on a single NVIDIA DGX Station with Ubuntu 20.04 LTS. The full hardware specification includes 8 Tesla V100 GPUs with 32GB VRAM, 512GB RAM, and Intel(R) Xeon(R) CPU E5-2698 v4. Each experiment was run using a single GPU. The code is based on the PyTorch \cite{NEURIPS2019_9015} framework.

\subsection{Architectures of the models}

\paragraph{StyleGAN backbone}
Our experiments were performed using the pre-trained, publicly available StyleGAN2 trained on the FFHQ dataset \cite{Karras2019stylegan2}.
\paragraph{\our{} for StyleGAN backbone} We use NICE architecture with $4$ coupling layers with $4$ layers in each and width $256$. We use Adam optimizer with learning rate $10^{-4}$ and train model for $1000$ epochs. The hyperparameters $\sigma_0$ and $\gamma$ used for modeling conditional distributions, are set to $0.4$ and $0.999$, respectively.

\paragraph{VAE backbone}
For our experiments, we reuse the VAE architecture from \cite{li2020latent}. We use an encoder with 5 convolutional layers starting with 128 filters and doubling. The decoder is symmetrical to the decoder. We use leakyReLU activations. We train the network for 50 epochs with batch size 40 and Adam optimizer with the learning rate set to $10^{-4}$. We additionally train a PatchGAN model \cite{li2016precomputed} to improve the sharpness of the images.

\paragraph{\our{} for VAE backbone}
As previously, we use NICE architecture with $4$ coupling layers with $4$ layers in each and width $256$. We train the model for $50$ epochs using Adam optimizer with learning rate $10^{-4}$ and $\sigma_0$. The hyperparameters $\gamma$ are set to $0.7$ and $0.99$, respectively.

\paragraph{cFlow}
We train cFlow model also on top of the base network. We use Conditional Masked Autoregressive Flow with $5$ layers of each consisting of reverse permutation and MADE component with $2$ residual blocks. Moreover, we have been encoding attributes using $1$ linear layer which was after that passed as a context input to the flow. We train the model for $50$ epochs using Adam optimizer witht learning rate $10^{-3}$. During sampling, the temperature trick was used with $T=0.7$.
 
\paragraph{ResNet classifier}
To evaluate the correctness of attribute manipulation in the case of CelebA dataset, we used a standard ResNet-56 classifier. We trained it on the task of multi-label classification, with class weighting to correct for class imbalance. We used the Adam optimizer with the learning rate set to $3 \cdot 10^{-4}$, batch size $256$ and trained it for $50$ epochs.

\subsection{Additional Results}

In this subsection, we present additional results and models comparison, which were not included in the main paper because of space restrictions.

\paragraph{Manipulating the StyleGAN latent codes}

In Figures \ref{fig:stylegan_interpolations} and \ref{fig:stylegan_modifications}, we present additional results of attribute manipulations performed by \our{} and StyleFlow on the latent codes of StyleGAN backbone. In most cases, \our{} modifies only the requested attribute leaving the remaining ones unchanged, which is not always the case of StyleFlow (compare 4th row of Figure \ref{fig:stylegan_interpolations} or 3rd row of Figure \ref{fig:stylegan_modifications}). This confirms that the latent space produced by \our{} is more disentangled than the one created by StyleFlow.

\paragraph{Manipulating images using VAE backbone}

In Figure \ref{fig:modifications2}, we show additional results of image manipulation performed by \our{}, MSP, and cFlow using VAE backbone. One can observe that \our{} and MSP perform the requested modification more accurately than cFlow. 

\paragraph{Manipulating attributes intensity of generated images.}

\begin{figure}[t!]
   \centering
    \includegraphics[width=\linewidth]{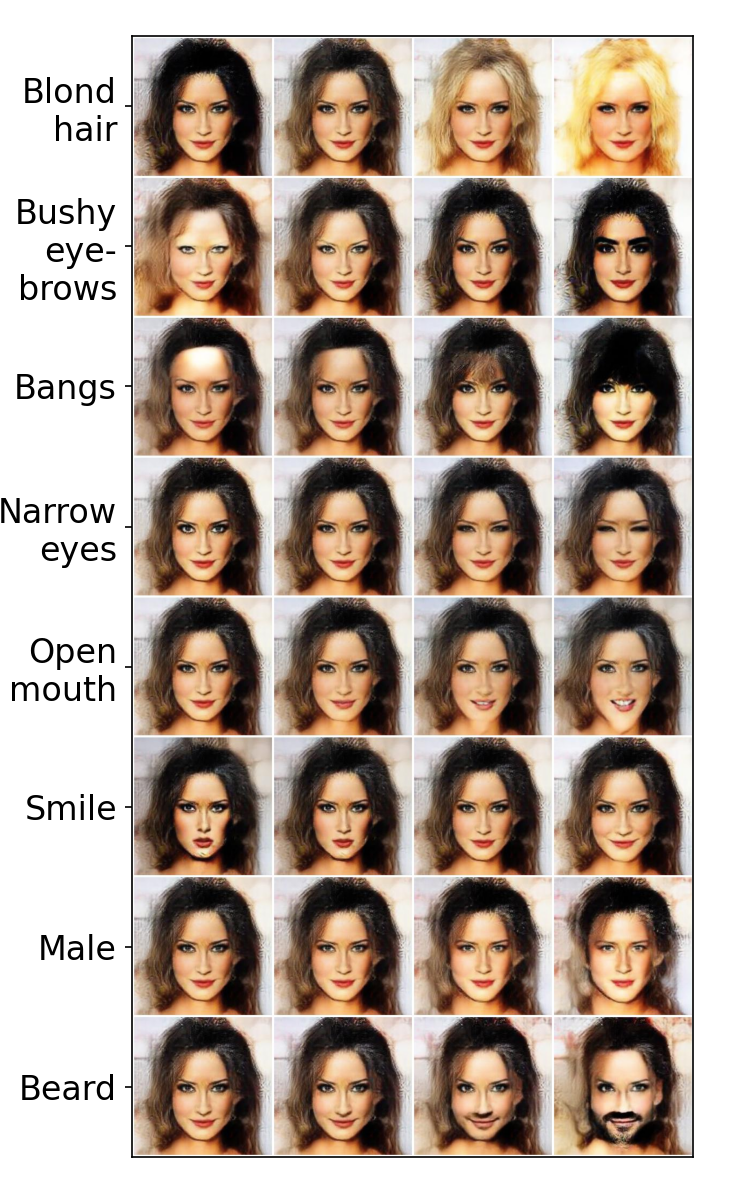}
    \caption{Manipulating the intensity of labeled attributes of the generated sample. Since \our{} models the values of the attributes with continuous distributions, it can control the intensity of each attribute and even sometimes extrapolate outside the data distribution (e.g. very bright blond hair). }
    \label{fig:generated_interpolations}
  \end{figure}

In this experiment, we consider images fully generated by \our{} (not reconstructed images) attached to the VAE backbone. More precisely, we sample a single style variable $\mathbf{s}$ from the prior distribution and manipulate the label variables of CelebA attributes. It is evident in Figure~\ref{fig:generated_interpolations} that \our{} freely interpolates between binary values of each attribute and even extrapolates outside the data distribution. This is possible thanks to the continuous form of prior distribution we are using in the latent space, which enables us to choose the intensity of each attribute. We emphasize that this information is not encoded in the dataset, where labels are represented as binary variables. However, in reality, an attribute such as 'narrow eyes' covers a whole spectrum of possible eyelid positions, from eyes fully closed, through half-closed to wide open. \our{} is able to recover this property without explicit supervision. Interestingly, we also see cases of extrapolation outside of the dataset, e.g. setting a significantly negative value of the 'bangs' attribute, which can be interpreted as an illogical condition 'extreme absence of bangs', creates a white spot on the forehead.

\begin{figure}
    \centering
    \includegraphics[width=\linewidth]{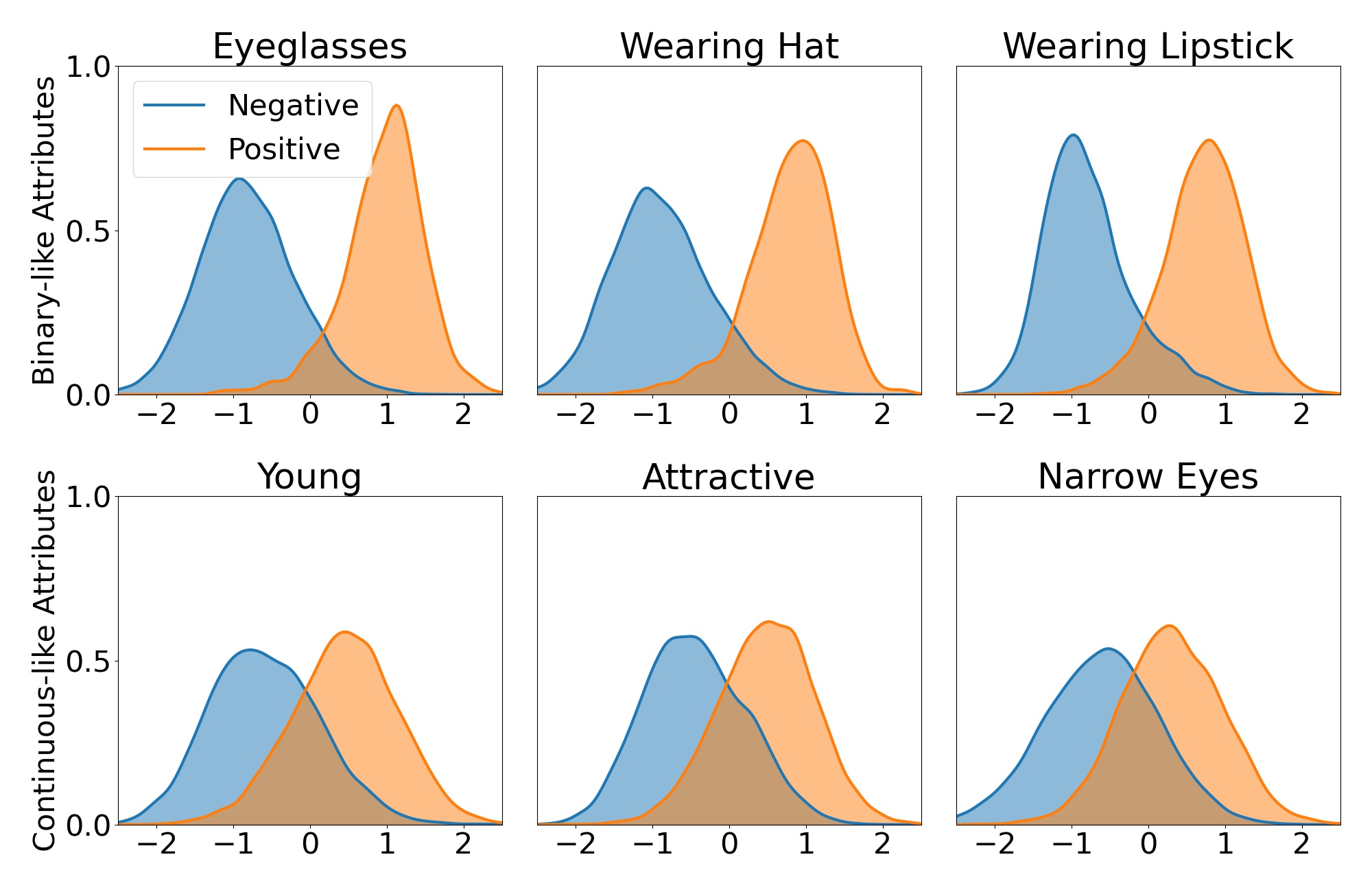}
    \caption{The density of the positive and negative samples for chosen attributes in the flow latent space estimated using all examples from the CelebA test set. Binary attributes (top row) are clearly separated while continuous attributes (bottom row) overlap significantly.}
    \label{fig:hist_attributes}
\end{figure}

Figure~\ref{fig:hist_attributes} shows that the shape of the empirical distributions in the latent space of \our{} allows for this continuous change. While the positive and negative classes of boolean attributes such as the presence of a hat or eyeglasses are clearly separated, in more continuous variables like youth and attractiveness they overlap significantly, allowing for smooth interpolations. This phenomenon emerges naturally, even though CelebA provides only binary labels for all the attributes.

\paragraph{Generation capabilities of MSP and the VAE backbone}

\begin{figure*}
    \centering
    {\bf \large VAE}\\
    \includegraphics[width=\textwidth]{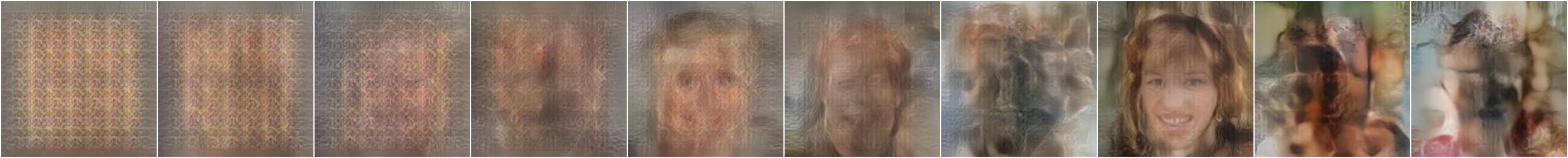}\\
    {\bf \large MSP}
    \includegraphics[width=\textwidth]{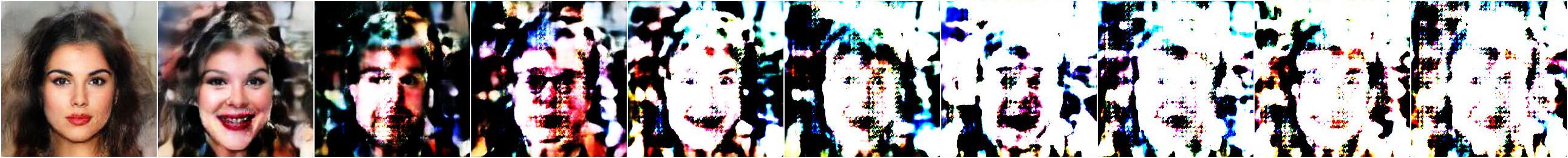}\\
    \begin{tabularx}{\linewidth}{CCCCCCCCCC} 0.1 & 0.2 & 0.3 & 0.4 & 0.5 & 0.6 & 0.7 & 0.8 & 0.9 & 1.0\end{tabularx}\\
    \caption{Samples from the base VAE (top row) and MSP (bottom row) models using increasing values of the temperature parameter (bottom line). MSP generates typical face images only for a very low temperature, while VAE does not generate face images at all.}
    \label{fig:vae}
\end{figure*}

In Figure \ref{fig:vae} (top), we demonstrate that the base VAE model taken from the MSP paper \cite{li2020latent} cannot generate new face images, but only manipulate the attributes of input examples. In consequence, it works similar to the autoencoder model. For this reason it is especially notable that \our{} can improve the generation performance of the backbone model (see the main paper). In contrast, MSP cannot generate new face images using this VAE model as shown in the bottom row of Figure \ref{fig:vae}. For very low temperatures, MSP generates typical (not diverse) faces.

\newcolumntype{C}{>{\centering\arraybackslash}X}

\paragraph{Generating images with attributes combinations taken from test set}

We present additional quantitative results for generating images with the requested combinations of attributes. In this experiment, we focus on typical combinations, which appear in a dataset. For this purpose, we generate 20,000 images with the same attribute combinations as in the CelebA test set. The results presented in Table \ref{tab:gen_metrics} show that \our{} outperforms both cFlow, cVAE, and $\Delta$-GAN in terms of classification scores.

\begin{table}[tb]
\caption{Average classification metrics for generating images with the combinations of attributes taken from the test set of CelebA.}
\label{tab:gen_metrics}
\centering
\begin{tabular}{lrrrr}
    \toprule
                &  
                \our{} &  cFlow & $\Delta$-GAN & cVAE   \\
    \midrule
        
        F1  
        & \textbf{0.69}   & 0.49     & 0.58        & 0.59   \\
        AUC  
        & \textbf{0.92}        & 0.85     & 0.87        & 0.88  \\
    \bottomrule
    \end{tabular}
\end{table}

\subsection{Ablations} \label{sec:cnf}

\paragraph{CNF vs NICE}

In our main experiments, we use the NICE \cite{dinh2014nice} approach to flow-based models. This choice was motivated by the computational and conceptual simplicity of the approach. However, we also empirically evaluated a more complex approach of continuous normalizing flows \cite{chen2018neural} which cast the distribution modeling task as a problem of solving differential equations. The CNF implementation consisted of $2$ stacked CNFs, each containing $3$ concatsquash layers with a hidden dimension $2048$. Table \ref{tab:cnf_class} shows the results of both approaches in the task of multi-label conditional generation using VAE backbone. We use the same combinations of attributes as in the CelebA test set. Table \ref{tab:cnf_independent} shows an analogical comparison when the attributes were sampled independently, which is more challenging setting. For both of these settings results on NICE and CNF are comparable. Although CNF samples get better FIDs, they also score worse on the classification metrics, which suggests that the model might be worse at enforcing the class conditions. Overall, both models perform similarly and because of that, we use NICE as the approach is less expensive computationally.

\begin{table}[t]
\centering
\caption{Average classification metrics and FIDs for generating images with the combinations of attributes taken from the test set of CelebA.}
\label{tab:cnf_class}
\begin{tabular}{lrrrr}
    \toprule
            & NICE    & CNF   \\
    \midrule
        FID & 72.72   & 68.96 \\
        F1  & 0.69    & 0.63  \\
        AUC & 0.92    & 0.89  \\
    \bottomrule
    \end{tabular}
\end{table}

\begin{table}[t]
\caption{Average classification metrics and FIDs for generating images, when the values of attributes were sampled independently.}
\label{tab:cnf_independent}
\centering
\begin{tabular}{lrrrr}
    \toprule
            & NICE     & CNF    \\
    \midrule
        FID & 77.48    & 73.31  \\
        F1  & 0.44     & 0.41   \\
        AUC & 0.78     & 0.75   \\
    \bottomrule
    \end{tabular}
\end{table}

\paragraph{Different autoencoder backbones}

In order to investigate how the structure of the latent space of the backbone autoencoder impacts the performance of our model, we check multiple $\beta$-VAE models with varying values of $\beta$. For each model we trained three architectures of INFs (small, medium, big) and picked the best performing ones for evaluation. The results presented in Table \ref{tab:backbone_results} show that the FID scores get worse as the value of $\beta$ increases. This is caused by the drop in the reconstructive power of the base model, which focuses more on the latent space regularization instead. Interestingly, the statistics also fall as the value of $\beta$ gets too low. The flow-based model cannot disentangle factors of variation from latent space which is not already at least partially structured. This experiment shows limitations of our model in respect to its reliance on the performance of the backbone autoencoder. However, \our{} is still quite robust as it achieves good results for a wide range of $\beta$ values. 

\begin{table}[t]
 \caption{Results for \our{} using $\beta$-VAE backbone for different values of $\beta$.}
    \label{tab:backbone_results}
    \small
    \centering
    \begin{tabular}{crrrrrr}
        \toprule
         $\beta$ & 0.5 & 1 & 2 & 4 & 8 & 16  \\
         \midrule
         FID & 61.86 & 55.11 & 61.96 & 65.76 & 77.94 & 110.46  \\
         F1 & 0.45 & 0.66 & 0.63 & 0.59 & 0.57 & 0.53  \\
         AUC &  0.79 & 0.90 & 0.88 & 0.87 & 0.86 & 0.83 \\
         \bottomrule
    \end{tabular}
\end{table}

\section{Details of molecules generation experiments}

\subsection{Background}

Designing a new drug is a long and expensive process that could cost up to 10 billion dollars and lasts even 10 years~\cite{mestre2012r}. The recent spread of SARS-CoV-2 virus and the pandemic it caused have shown how important it is to speed up this process. Recently, deep learning is gaining popularity in the cheminformatics community, where it is used to propose new drug candidates. However, using neural networks in the drug generation task is not easy and is fraught with problems. The complexity of the chemical space is high and thus training generative and predictive models is challenging. Although there are around $10^{60}$ of possible molecules~\cite{bohacek1996art}, detailed information (such as class labels) is known only about a small percentage of them. For example, the ChEMBL database~\cite{gaulton2017chembl}, one of the biggest databases with information about the molecular attributes, contains data for 2.1 M chemical compounds. Moreover, since obtaining labeled data requires long and costly laboratory experiments, the amount of labeled molecules in the training datasets is usually really small (often less than 1000), which is often not sufficient to train a good model. This poses an important research problem.

Deep neural networks are mostly used in cheminformatics for the following tasks:
\begin{itemize}
    \item virtual screening -- the search for potentially active compounds in the libraries of commercially available molecules using predictive models~\cite{coley2017convolutional,yang2019analyzing},
    \item de novo design -- generating new compounds with desirable properties that are not present in the above-mentioned libraries~\cite{olivecrona2017molecular,popova2018deep},
    \item lead optimization -- improving selected promising compounds to meet certain criteria~\cite{jin2018learning,maziarka2020mol}.
\end{itemize}

\our{} can be used for the two latter tasks, as our model can generate molecules with specified values of given attributes as well as optimize molecules by changing the value of selected labels.

\paragraph{SMILES representation}
SMILES~\cite{weininger1988smiles} (simplified molecular-input line-entry system) is a notation, for describing the structure of chemical species using a sequence of characters. SMILES representation consists of a specially defined grammar, which guarantees that a correct SMILES defines a unique molecule. The opposite is not actually true, as a molecule could be encoded by multiple SMILES representations. In order to add this property, the community introduced the canonicalization algorithm, which returns the canonical SMILES that is unique for each molecule.

\begin{figure}[t]
    \centering
    \subfloat[Melatonin]{
        \includegraphics[width=0.65\linewidth]{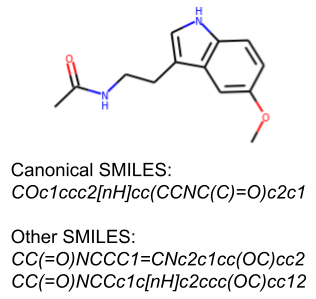}
    }
    \newline
    \subfloat[Vanillin]{
        \includegraphics[width=0.65\linewidth]{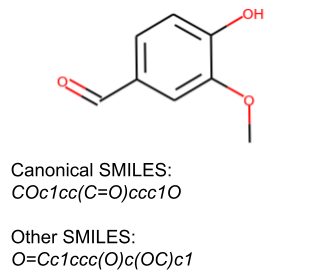}
    }\\
    \caption{Sample molecules together with their SMILES representations.}
    \label{fig:smiles_representation}
\end{figure}

In Figure~\ref{fig:smiles_representation} we show two molecules together with their canonical SMILES as well as other SMILES representations.

\begin{figure}[t]
    \centering
    \includegraphics[width=\linewidth]{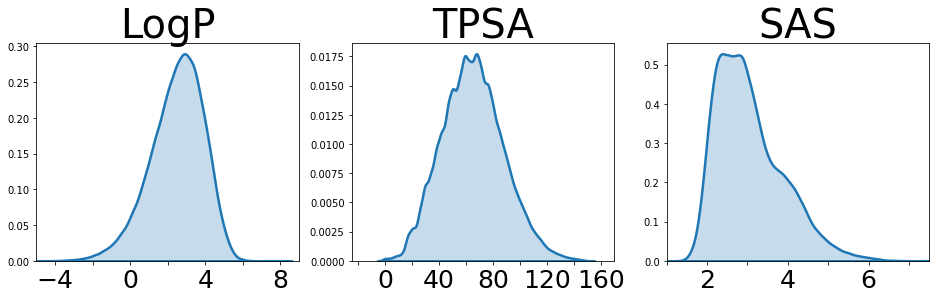}
    \caption{Density plots of chemistry attributes present in the training dataset.}
    \label{fig:chemistry_dataset_density}
\end{figure}

\begin{table}[t]
\caption{Correlations of attributes for chemical molecules modeling.}
    \label{tab:chemistry_corr}
    \small
    \centering
    \begin{tabular}{crrrrrr}
        \toprule
         {} & logP & TPSA & SAS \\
         \midrule
         logP & 1.00 & -0.16 & 0.51 \\
         TPSA & -0.16 & 1.00 & -0.18 \\
         SAS  & -0.51 & -0.18 & 1.00 \\
         \bottomrule
    \end{tabular}
\end{table}

\paragraph{Modeled attributes}

In our chemistry experiments, we modeled 3 chemical attributes: logP, TPSA, and SAS. Below, we describe their responsibilities:

\begin{itemize}
    \item logP -- logarithm of the partition coefficient. Describes the molecule solubility in fats. It shows how well the molecule is passing through membranes.
    \item TPSA -- the topological polar surface area of a molecule is the surface sum over all polar atoms or molecules (together with their attached hydrogen atoms). TPSA could be used as a metric of the ability of a drug to permeate cells.
    \item SAS -- synthetic accessibility score defines the ease of synthesis of a drug-like molecule. When generating a drug candidate, one would rather want it to be easily synthesized so that it can be obtained in the laboratory.
\end{itemize}

\paragraph{Dataset}

We conducted our chemistry experiments using a dataset of 250k molecules sampled from the ZINC database~\cite{sterling2015zinc}, which is a dataset of commercially available chemical compounds. The mean number of SMILES tokens in our dataset is equal to 38.31, with a standard deviation equal to 8.46.

Figure~\ref{fig:chemistry_dataset_density} shows the distribution of attributes of molecules that make up our training dataset.

Since the values of the chemical attributes are related to the structure of the molecule, many of them will be correlated in some way. In Table~\ref{tab:chemistry_corr} we present the correlations between the chemistry attributes. The correlations suggest that it might be difficult or even impossible to manipulate logP and SAS attributes independently, setting a difficult challenge for \our{}.

\subsection{Hyperparameters}

\paragraph{VAE}
The encoder consists of 3 bi-GRU~\cite{cho2014learning} layers, with hidden size equal to 256 and output size (latent dimensionality) equal to 100. The decoder consists of 3 GRU layers with the hidden size equal to 256. The architecture of the backbone model is significantly different from the one used in the image domain, which partially confirms that \our{} can be combined with various autoencoder models.

We trained the VAE model for 100 epochs, using batch size of 256 and learning rate equal to 1e-4.

\paragraph{NICE}
The flow model consisted of 6 coupling layers, each of which consists of 6 dense layers with a hidden size equal to 256. We trained NICE for 50 epochs, with learning rate equal to 1e-4 and batch size 256. We used $\sigma_0 = 1.0$ and $\gamma=0.9$.

\begin{figure}[t]
    \centering
    \subfloat[LogP = 1.0, TPSA = 60.0, SAS = 5.0]{
        \includegraphics[width=\linewidth]{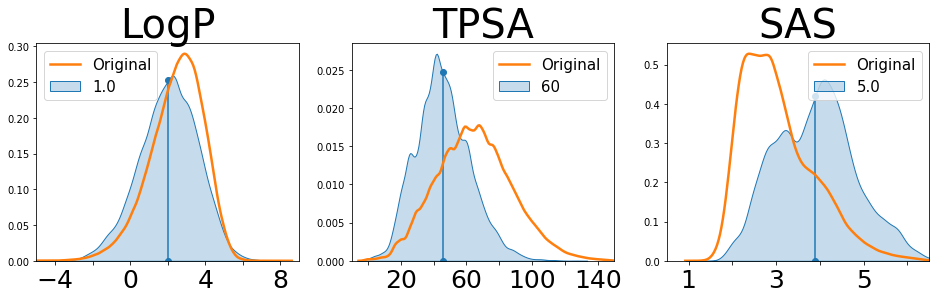}
    }
    \newline
    \subfloat[LogP = 3.0, TPSA = 75.0, SAS = 3.0]{
        \includegraphics[width=\linewidth]{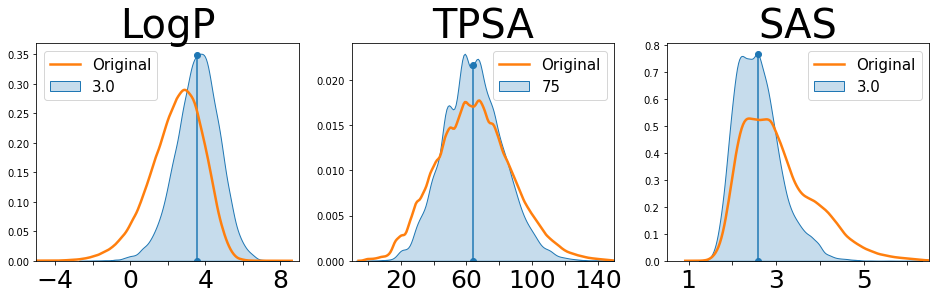}
    }
    \newline
    \subfloat[LogP = 5.0, TPSA = 50.0, SAS = 2.0]{
        \includegraphics[width=\linewidth]{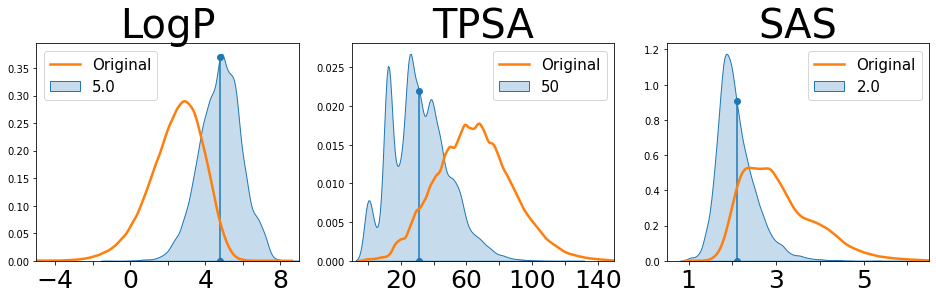}
    }
    \caption{Distribution of labeled attributes for generated molecules (for the experiment with multiple attributes condition), together with distribution for the training dataset. The average of every distribution is marked with a vertical line.}
    \label{fig:conditional_generation_app}
\end{figure}

\begin{figure}[t]
    \centering
    \subfloat[Molecules decoded from path]{
        \includegraphics[width=0.45\linewidth]{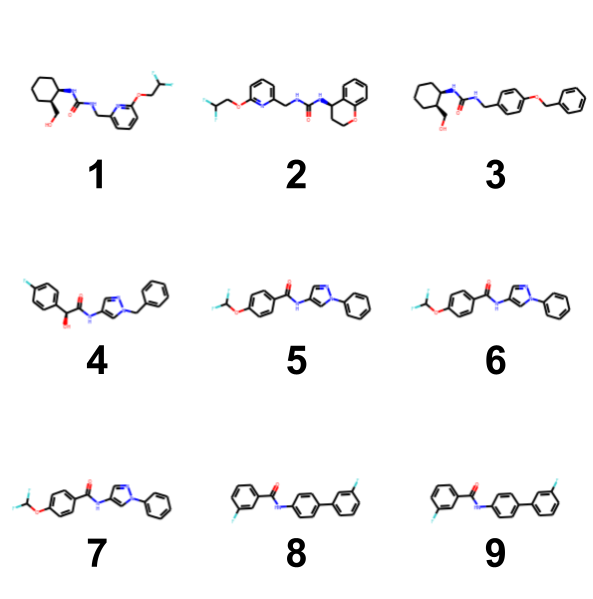}
    }
    \subfloat[LogP of presented molecules]{
        \includegraphics[width=0.45\linewidth]{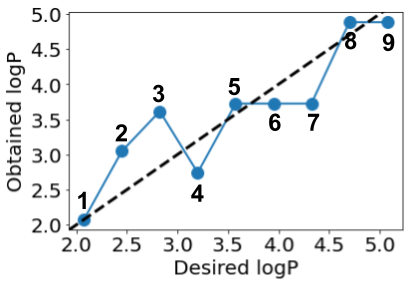}
    }\\
    \caption{Molecules obtained by the model during an optimization phase (left side), together with their LogP (right side).}
    \label{fig:traversal_chemistry_app_logP_1}
\end{figure}

\begin{figure}[t]
    \centering
    \subfloat[Molecules decoded from path]{
        \includegraphics[width=0.45\linewidth]{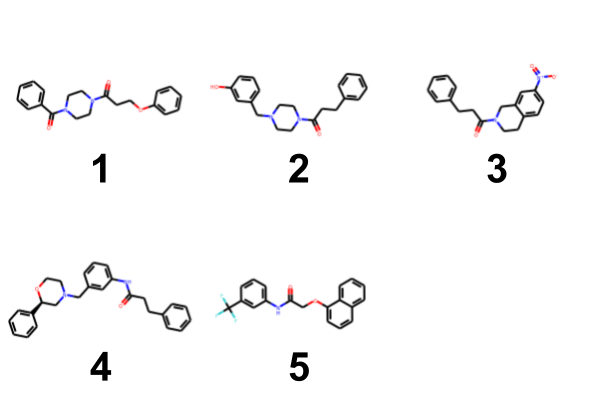}
    }
    \subfloat[LogP of presented molecules]{
        \includegraphics[width=0.45\linewidth]{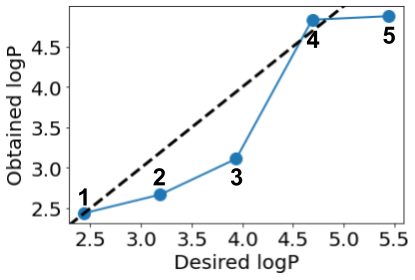}
    }\\
    \caption{Molecules obtained by the model during an optimization phase (left side), together with their LogP (right side).}
    \label{fig:traversal_chemistry_app_logP_2}
\end{figure}

\begin{figure}[t]
    \centering
    \subfloat[Molecules decoded from path]{
        \includegraphics[width=0.45\linewidth]{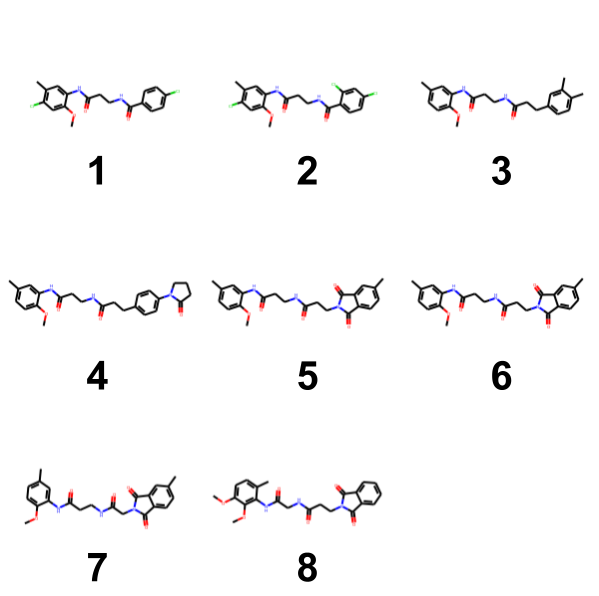}
    }
    \subfloat[TPSA of presented molecules]{
        \includegraphics[width=0.45\linewidth]{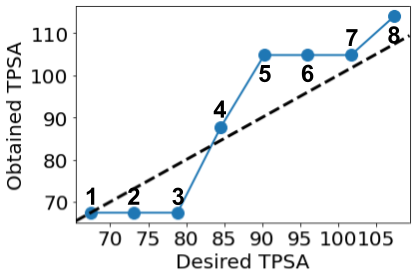}
    }\\
    \caption{Molecules obtained by the model during an optimization phase (left side), together with their TPSA (right side).}
    \label{fig:traversal_chemistry_app_TPSA_1}
\end{figure}

\begin{figure}[t]
    \centering
    \subfloat[Molecules decoded from path]{
        \includegraphics[width=0.45\linewidth]{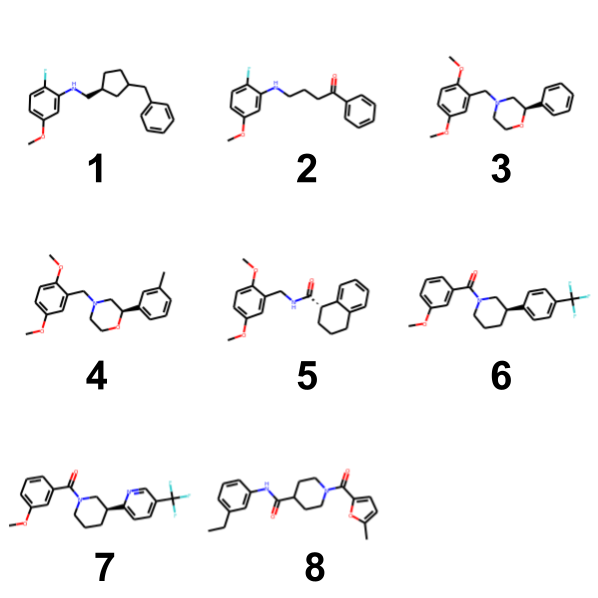}
    }
    \subfloat[TPSA of presented molecules]{
        \includegraphics[width=0.45\linewidth]{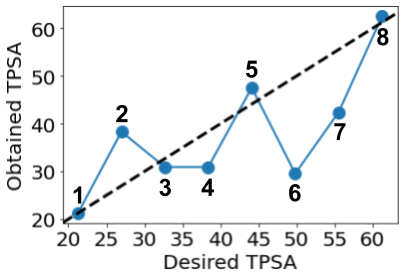}
    }\\
    \caption{Molecules obtained by the model during an optimization phase (left side), together with their TPSA (right side).}
    \label{fig:traversal_chemistry_app_TPSA_2}
\end{figure}

\begin{figure}[t]
    \centering
    \subfloat[Molecules decoded from path]{
        \includegraphics[width=0.45\linewidth]{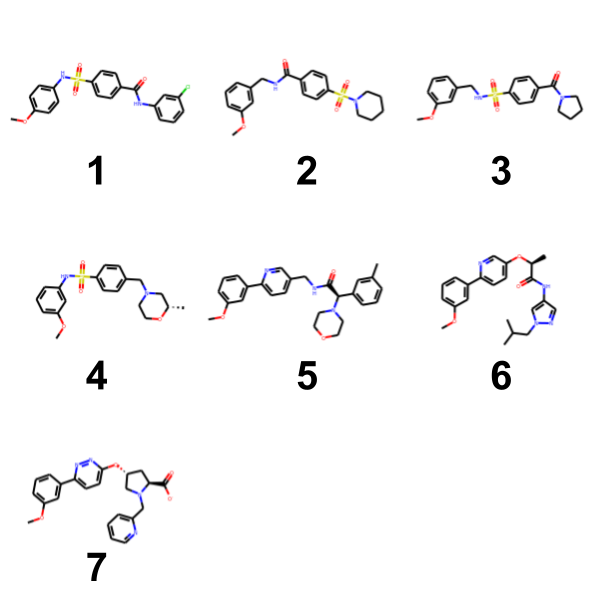}
    }
    \subfloat[SAS of presented molecules]{
        \includegraphics[width=0.45\linewidth]{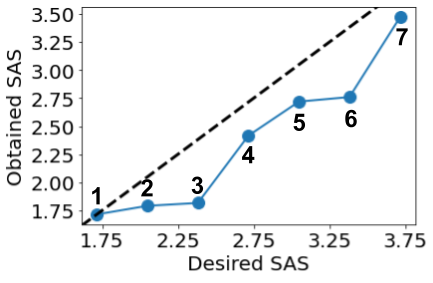}
    }\\
    \caption{Molecules obtained by the model during an optimization phase (left side), together with their SAS (right side).}
    \label{fig:traversal_chemistry_app_SAS_1}
\end{figure}

\begin{figure}[t]
    \centering
    \subfloat[Molecules decoded from path]{
        \includegraphics[width=0.45\linewidth]{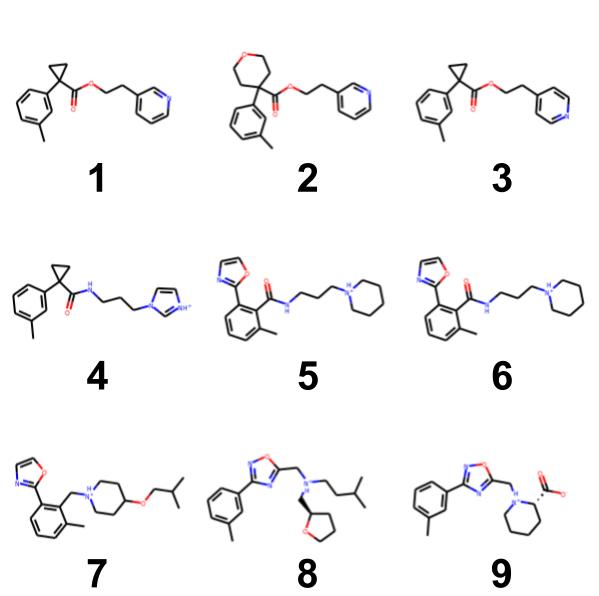}
    }
    \subfloat[SAS of presented molecules]{
        \includegraphics[width=0.45\linewidth]{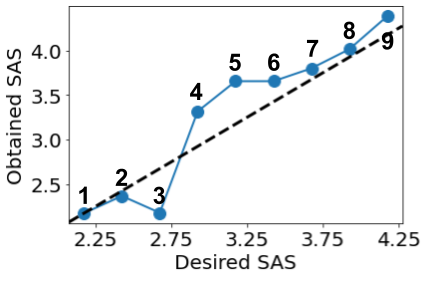}
    }\\
    \caption{Molecules obtained by the model during an optimization phase (left side), together with their SAS (right side).}
    \label{fig:traversal_chemistry_app_SAS_2}
\end{figure}

\subsection{Additional experiments}

In the following subsection, we show additional results for the chemistry-based experiments, for both conditional generation as well as latent space traversal. Furthermore, we show how \our{} works with the conditional normalizing flow instead of NICE as a base flow model.

\paragraph{Conditional generation}

In the main paper, we presented results for conditional generation in the setting of a single attribute condition (where the value of the remaining attributes was sampled from their prior distribution). Here we also show results for a situation where set conditions on all attributes at the same time. 

In particular, we tested 3 different settings:
\begin{enumerate}
    \item LogP set to 1.0, TPSA set to 60.0, SAS set to 5.0. 
    \item LogP set to 3.0, TPSA set to 75.0, SAS set to 3.0.
    \item LogP set to 5.0, TPSA set to 50.0, SAS set to 2.0. 
\end{enumerate}

The density plots of the attributes of the molecules generated in these settings are presented in Figure~\ref{fig:conditional_generation_app}.

\paragraph{Latent space traversal}

We also present more results for latent space traversals, which is a task that imitates the inter-class interpolation experiments from the image domain. 
For this purpose, we tested how \our{} can traverse the latent space of CharVAE. Therefore, we selected a few random molecules from our dataset, and for every one, we forced \our{} 
to gradually increase the value of the specified attribute by some value and decoded the resulting molecules back into the latent space. 
The goal of this task is to generate the molecules that are structurally similar to the initial one, except for changes in the desired attributes. This is an important challenge in the \textit{lead optimization stage} of the drug discovery process.

\textbf{LogP} For LogP, we forced \our{} to increase the molecular attribute value by 3.
Figures~\ref{fig:traversal_chemistry_app_logP_1} and \ref{fig:traversal_chemistry_app_logP_2} show the obtained molecules, together with the optimized attribute values.

\textbf{TPSA} For TPSA, we forced \our{} to increase the molecular attribute value by 40.
Figures~\ref{fig:traversal_chemistry_app_TPSA_1} and \ref{fig:traversal_chemistry_app_TPSA_2} show the obtained molecules, together with the optimized attribute values.

\textbf{SAS.} For SAS, we forced \our{} to increase the molecular attribute value by 2.
Figures~\ref{fig:traversal_chemistry_app_SAS_1} and \ref{fig:traversal_chemistry_app_SAS_2} show the obtained molecules, together with the optimized attribute values.

\paragraph{CNF vs NICE}

\begin{figure}[t]
    \centering
    \includegraphics[width=\linewidth]{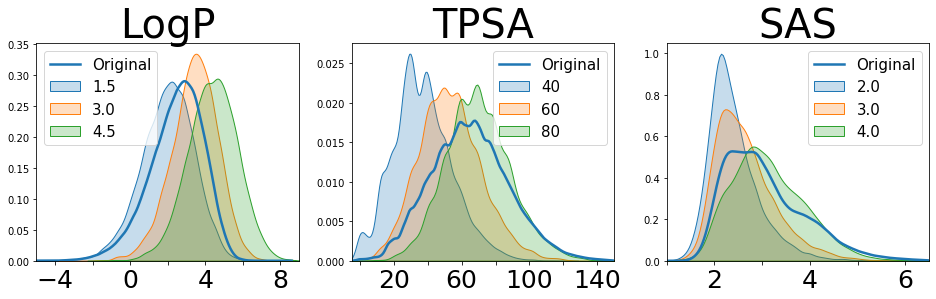}
    \caption{Distribution of labeled attributes for generated molecules for \our{} with the conditional normalizing flow, together with distribution for the training dataset. Each color shows the value of the labeled attribute that was used for generation.}
    \label{fig:conditional_generation_cnf}
\end{figure}

We also tested how replacing NICE~\cite{dinh2014nice} with conditional normalizing flow~\cite{chen2018neural} affects the process of molecular generation using \our{}. For this purpose, we repeated the chemistry-based conditional generation experiments from the main text, but with CNF as our backbone flow model. Results are presented in Figure~\ref{fig:conditional_generation_cnf}. One can see, that in this version \our{} is also capable of moving the density of the attributes of the generated molecules towards the desired value. The obtained changes, however, are worse than in the case of NICE as a flow backbone.

\end{document}